%% file: main.tex
\documentclass[12pt]{article}

\usepackage{geometry}
\usepackage{package/tech-report}
\usepackage{import}
\usepackage{array}
\usepackage{amsmath}
\usepackage{amssymb}
\usepackage{booktabs}
\usepackage{graphicx}
\usepackage{hyperref}
\usepackage{caption}
\usepackage{enumitem}
\usepackage{xcolor}
\usepackage{caption}
\usepackage{subcaption}
\usepackage{upgreek,textgreek}
\usepackage{tikz,lipsum,lmodern}
\usepackage[most]{tcolorbox}
\usepackage{listings}
\usepackage{cleveref}
\usepackage{graphicx}
\usepackage[flushleft]{threeparttable}
\usepackage{titletoc}
\usepackage{upquote}
\usepackage{minted}
\usepackage{xcolor} 
\usepackage{algorithm} 
\usepackage{float} 
\usepackage[T1]{fontenc}
\definecolor{lightgray}{rgb}{0.95,0.95,0.95}

\input{package/math_command.tex}

\begin{document}

\title{Home-made Diffusion Model from Scratch to Hatch}

\newcommand{\name}{\fontsize{12pt}{14pt}\selectfont}
\newcommand{\aff}{\fontsize{12pt}{14pt}\selectfont}
\newcommand{\email}{\fontfamily{lmtt}\fontseries{l}\fontsize{9pt}{10pt}\selectfont}

\author{
  \name Shih-Ying Yeh \\
  \aff National Tsing Hua University \\
  \email kohaku@kblueleaf.net \\[1em]
  Code: \url{https://github.com/KohakuBlueleaf/HDM} \\
  Model: \url{https://huggingface.co/KBlueLeaf/HDM-xut-340M-anime}
}

\maketitle

\import{sections/}{abstract.tex}
\import{sections/}{introduction.tex}
\import{sections/}{preliminary.tex}
\import{sections/}{method.tex}
\import{sections/}{training.tex}
\import{sections/}{result.tex}
\import{sections/}{conclusion.tex}

\newpage
\bibliographystyle{plainnat}
\bibliography{reference}

\import{sections/}{appendix.tex}

\end{document}

%% file: package/math_command.tex
\usepackage{amsmath,amsfonts,bm}

\def\eqref#1{equation~\ref{#1}}

\def\1{\bm{1}}

\DeclareMathAlphabet{\mathsfit}{\encodingdefault}{\sfdefault}{m}{sl}
\SetMathAlphabet{\mathsfit}{bold}{\encodingdefault}{\sfdefault}{bx}{n}



%% file: sections/abstract.tex
\begin{abstract}
We introduce Home-made Diffusion Model (HDM), an efficient yet powerful text-to-image diffusion model optimized for training (and inferring) on consumer-grade hardware. HDM achieves competitive 1024×1024 generation quality while maintaining a remarkably low training cost of \$535-620 using four RTX5090 GPUs, representing a significant reduction in computational requirements compared to traditional approaches. Our key contributions include: (1) \textbf{Cross-U-Transformer (XUT)}, a novel U-shape transformer, Cross-U-Transformer (XUT), that employs cross-attention for skip connections, providing superior feature integration that leads to remarkable compositional consistency; (2) \textbf{a comprehensive training recipe} that incorporates TREAD acceleration, a novel shifted square crop strategy for efficient arbitrary aspect-ratio training, and progressive resolution scaling; and (3) \textbf{an empirical demonstration} that smaller models (343M parameters) with carefully crafted architectures can achieve high-quality results and emergent capabilities, such as intuitive camera control. Our work provides an alternative paradigm of scaling, demonstrating a viable path toward democratizing high-quality text-to-image generation for individual researchers and smaller organizations with limited computational resources.
\end{abstract}

%% file: sections/introduction.tex
\section{Introduction}

Text-to-image (T2I) generation represents one of the most important and challenging tasks in generative modeling. Recent advances in the field have been dominated by a scaling paradigm, where ever-larger models are trained on massive datasets to achieve state-of-the-art performance~\citep{podell2024sdxl, esser2024scalingrectifiedflowtransformers, labs2025flux1kontextflowmatching, wu2025qwenimagetechnicalreport}. While this approach has yielded impressive results, it has also created significant barriers to entry, with training costs often exceeding tens of thousands of dollars on expensive data center infrastructure. 

This trend toward increasingly compute-intensive approaches—characterized by larger model architectures and extended training regimens—has created an accessibility gap, limiting innovation to well-resourced institutions and hindering broader participation in AI research and development.

\textbf{Our Approach.} We argue that architectural innovation and training efficiency can serve as a powerful alternative to pure scaling. We introduce the Home-made Diffusion Model (HDM), an effective T2I model designed from the ground up to be trained entirely on consumer-level hardware. Our approach centers on an efficiency-first philosophy, building upon the foundations of U-shaped transformer architectures like UViT~\citep{bao2023worthwordsvitbackbone} and HDiT~\citep{crowson2024scalable} but introducing novel mechanisms that enhance performance and capabilities within a constrained computational budget. Rather than simply scaling down existing models, we focus on targeted optimizations that democratize access to high-quality T2I generation while unlocking unique and practical functionalities.

\textbf{Key Contributions.} We successfully demonstrate the viability of training a competitive T2I model at home, hence the name Home-made Diffusion Model. Our specific contributions include:

\begin{itemize}
\item \textbf{Cross-U-Transformer (XUT):} A novel U-shaped transformer architecture that replaces traditional concatenation-based skip connections with cross-attention mechanisms. This design enables more sophisticated feature integration between encoder and decoder layers, leading to remarkable compositional consistency across prompt variations.

\item \textbf{Comprehensive Training Recipe:} A complete and replicable training methodology incorporating TREAD acceleration for faster convergence~\citep{krause2025treadtokenroutingefficient}, a novel \textit{Shifted Square Crop} strategy that enables efficient arbitrary aspect-ratio training without complex data bucketing, and progressive resolution scaling from 256² to 1024².

\item \textbf{Empirical Demonstration of Efficient Scaling:} We demonstrate that smaller models (343M parameters) with carefully crafted architectures can achieve high-quality 1024×1024 generation results while being trainable for under \$620 on consumer hardware (four RTX5090 GPUs). This approach reduces financial barriers by an order of magnitude and reveals emergent capabilities such as intuitive camera control through position map manipulation—capabilities that arise naturally from our training strategy without additional conditioning.
\end{itemize}

%% file: sections/preliminary.tex
\section{Preliminaries}
\label{sec:preliminaries}

This section provides essential background on diffusion models, text-guided generation, and transformer architectures that form the foundation of our approach.

\subsection{Diffusion Models and Flow Matching}

Diffusion models have emerged as the dominant paradigm for high-quality image generation due to their stable training dynamics and superior sample quality. The foundational DDPM framework ~\citep{10.5555/3045118.3045358, 10.5555/3495724.3496298} establishes a forward noising process that gradually corrupts data with Gaussian noise:
\begin{align}
q(x_t|x_{t-1}) = \mathcal{N}(x_t; \sqrt{1-\alpha_t}x_{t-1}, \alpha_t \mathbf{I})
\end{align}
and a learned reverse process that generates samples by iteratively denoising:
\begin{align}
p_\theta(x_{t-1}|x_t) = \mathcal{N}(x_{t-1}; \mu_\theta(x_t, t), \Sigma_\theta(x_t, t))
\end{align}

More recently, flow matching approaches ~\citep{lipman2023flow} provide an alternative formulation with improved training dynamics and sampling efficiency. Flow matching defines a continuous-time generative process through ordinary differential equations (ODEs):
\begin{align}
\frac{dx}{dt} = v_t(x)
\end{align}
where $v_t(x)$ is a learned vector field. The flow matching objective minimizes:
\begin{align}
\mathcal{L}_{\text{FM}} = \mathbb{E}_{t,x_0,x_1}\left[\|v_\theta(x_t, t) - (x_1 - x_0)\|^2\right]
\end{align}

This formulation can achieve similar generation quality with potentially improved training dynamics and more efficient sampling procedures, making it particularly suitable for resource-constrained training scenarios.

\subsection{Text-Guided Diffusion Models}

The evolution of text encoders in diffusion models has progressed through several distinct generations, each bringing improvements in text understanding and generation quality. Early models like Stable Diffusion relied primarily on CLIP ~\citep{radford2021learningtransferablevisualmodels} text encoders  ~\citep{rombach2021highresolution, podell2024sdxl, esser2024scalingrectifiedflowtransformers}, which provided basic text-image alignment capabilities but limited semantic understanding cite.

More recent approaches like SD3/SD3.5 incorporate T5-XXL encoders ~\citep{esser2024scalingrectifiedflowtransformers, flux2024} for enhanced text understanding, significantly improving prompt adherence and complex scene composition. A notable emerging trend involves using causal language models as text encoders, with models like LLaMA, Gemma, and Qwen demonstrating effectiveness in this role ~\citep{wu2025qwenimagetechnicalreport, qin2025luminaimage20unifiedefficient}.

While Qwen has specifically released embedding models based on their language model architectures, we maintain the causal language model structure and utilize the final hidden state as our text embedding, leveraging the rich contextual representations learned during language modeling pretraining.

\subsection{U-Shaped Transformer Architectures}

Standard Diffusion Transformers (DiT) ~\citep{Peebles2022DiT} prove insufficient for complex image generation tasks due to their lack of multi-scale feature learning capabilities. Following UNet design principles, we adopt U-shaped transformer architectures that combine the benefits of hierarchical feature learning with transformer scalability.

UViT and HDiT represent two distinct philosophical approaches to U-shaped transformers:

\textbf{UViT Architecture:} ~\citep{bao2023worthwordsvitbackbone} maintains consistent resolution throughout the network, relying solely on U-shaped skip connections to facilitate multi-scale feature learning. UViT proposes that all conditioning information (including timesteps, class conditions, text conditions) should be treated as input tokens rather than using adaptive layer normalization (adaLN) or cross-attention. However, practical implementations like HunYuan-DiT, an open-source T2I model utilizing UViT architecture, still employ adaLN and cross-attention for text conditioning.

\textbf{HDiT Architecture:} ~\citep{crowson2024scalable} incorporates explicit downsampling/upsampling operations and employs local attention mechanisms (NATTEN) in higher-resolution layers while using global attention in lower-resolution layers. This approach provides significant computational speedup compared to standard DiT while maintaining a more UNet-like intuition for multi-scale feature learning.

\subsection{Efficient Training Strategies}

Recent advances in efficient diffusion training have introduced several notable approaches addressing the computational challenges of large-scale training. SANA ~\citep{xie2025sana} achieves dramatic sequence length reduction through DC-AE compression ~\citep{chen2025deep, chen2025dcae15acceleratingdiffusion} and employs linear attention mechanisms to avoid quadratic complexity, though with some notable quality trade-offs in fine detail generation.

REPA/REPA-E ~\citep{yu2025representation, leng2025repaeunlockingvaeendtoend} focus on convergence acceleration through reference-enhanced progressive augmentation but are not specifically designed for extremely constrained training budgets, requiring additional computational overhead for reference feature generation.

TREAD ~\citep{krause2025treadtokenroutingefficient} provides both convergence acceleration and step speed improvements through architecture-agnostic token routing, making it particularly suitable for resource-constrained scenarios. Unlike alternatives that require auxiliary models or additional computational overhead, TREAD achieves efficiency gains without external dependencies, making it ideal for consumer-level training environments.

%% file: sections/method.tex
\section{HDM Methodology}
\label{sec:methodology}

\begin{figure*}[h]
\centering
\includegraphics[width=0.9\textwidth]{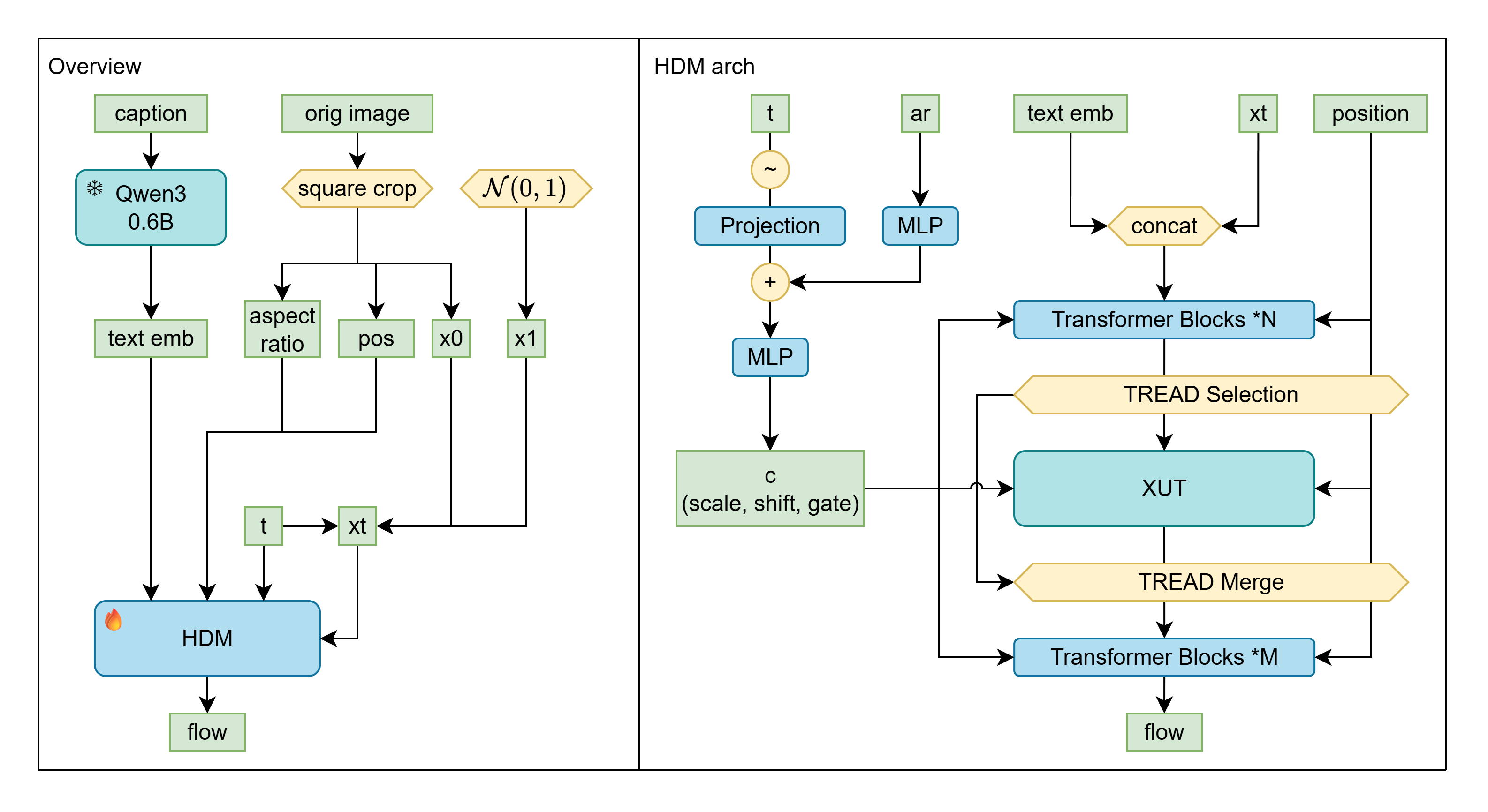}
\caption{HDM system overview showing the complete pipeline from input processing through XUT backbone to final output. The architecture integrates text encoding via Qwen3-0.6B, shifted square crop processing, and our novel XUT backbone with TREAD acceleration.}
\label{fig:hdm_overview}
\end{figure*}

This section presents our HDM approach, beginning with the novel XUT architecture and proceeding through our comprehensive training methodology. Figure~\ref{fig:hdm_overview} provides a system overview of the complete HDM pipeline.

\subsection{XUT (Cross-U-Transformer) Architecture}

Our XUT architecture draws inspiration from UViT/HDiT while introducing a novel approach to skip connections that addresses fundamental limitations in existing U-shaped transformer designs. We conceptualize the U-shaped architecture through the abstract pattern: \texttt{encoder → middle → decoder} with sophisticated feature merging mechanisms.

Traditional U-shaped implementations typically use concatenation or element-wise addition for skip connection operations, which can lead to suboptimal feature integration and information bottlenecks. In contrast, XUT employs cross-attention mechanisms to perform feature merging, leading us to term this approach the Cross-U-Transformer (XUT).

\textbf{Architecture Design.} The architecture repeats the encoder-decoder pattern $n_{\text{depth}}$ times, with each depth level containing $n_{\text{enc}}$ transformer blocks in the encoder path and $n_{\text{dec}}$ transformer blocks in the decoder path. This results in $(n_{\text{enc}} + n_{\text{dec}}) \times n_{\text{depth}}$ transformer blocks in total, plus $n_{\text{depth}}$ cross-attention blocks strategically placed in the first decoder transformer block of each depth level.

Formally, for depth level $d \in \{1, 2, \ldots, n_{\text{depth}}\}$, we define the forward pass as:
\begin{align}
h_d^{\text{enc}} &= \text{EncoderBlocks}^{(d)}(h_{d-1}^{\text{out}}) \label{eq:encoder}\\
h_d^{\text{dec}} &= \text{CrossAttn}(h_d^{\text{enc}}, h_{n_{\text{depth}}-d+1}^{\text{enc}}) \nonumber\\
&\quad + \text{DecoderBlocks}^{(d)}(h_d^{\text{enc}}) \label{eq:decoder}
\end{align}
where $h_0^{\text{out}}$ represents the input features, and the cross-attention mechanism enables selective information transfer from corresponding encoder depths through the symmetric indexing $h_{n_{\text{depth}}-d+1}^{\text{enc}}$.

\textbf{Concrete Example.} To illustrate the architecture, consider a configuration with $n_{\text{enc}}=1$, $n_{\text{dec}}=2$, $n_{\text{depth}}=2$. The forward pass follows:
\begin{align}
&\text{input} \xrightarrow{\text{trns}} a \xrightarrow{\text{trns}} b \xrightarrow{\text{trns}(b,b)} c \nonumber\\
&\quad \xrightarrow{\text{trns}} c' \xrightarrow{\text{trns}(c',a)} d \xrightarrow{\text{trns}} \text{output}
\end{align}
where $a, b, c, d$ represent intermediate hidden states, \texttt{trns} denotes transformer blocks, and \texttt{trns}$(q, kv)$ indicates cross-attention with query $q$ and key-value pairs from $kv$.

\begin{figure}[h]
\centering
\includegraphics[width=0.99\textwidth]{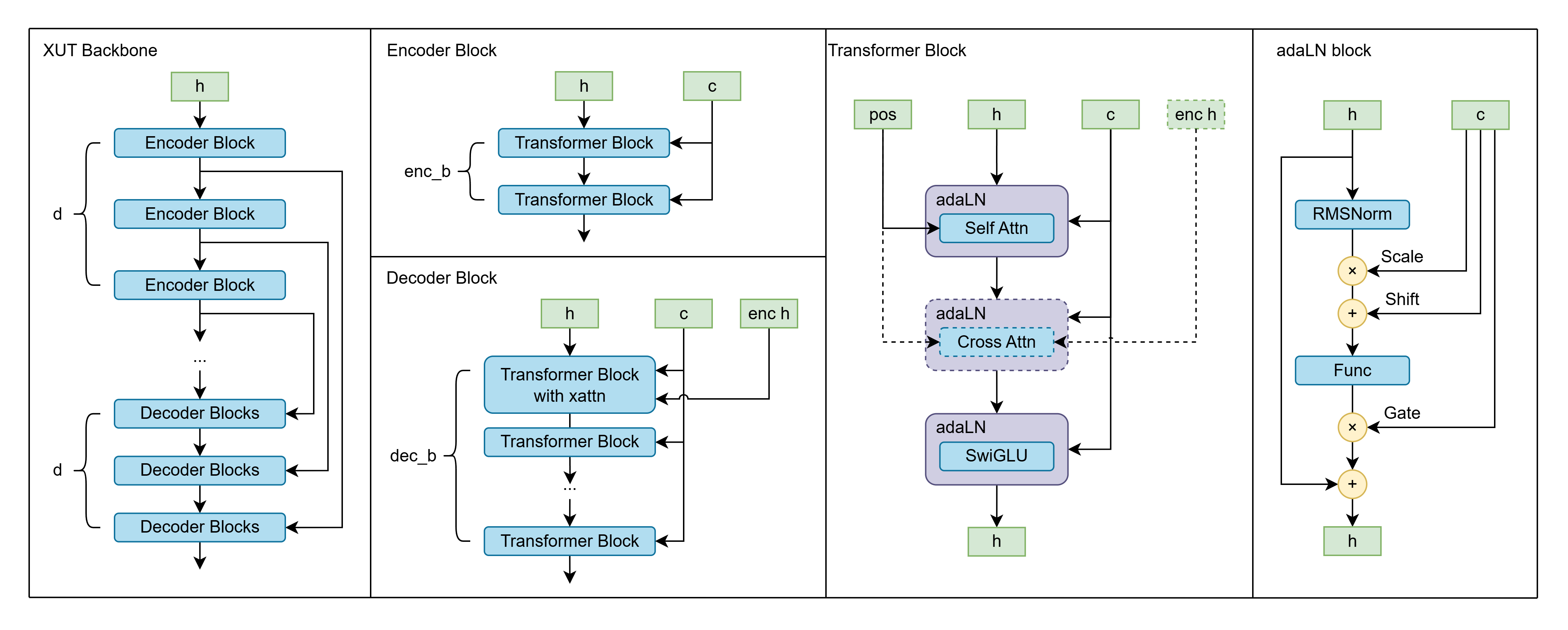}
\caption{Detailed XUT backbone architecture showing the encoder-decoder structure with cross-attention skip connections. Cross-attention mechanisms replace traditional concatenation-based skip connections, enabling sophisticated feature integration between encoder and decoder layers.}
\label{fig:xut_arch}
\end{figure}

\textbf{Rationale for Cross-Attention.} U-shaped skip connections facilitate multi-scale feature learning by transmitting early-stage features to corresponding later-stage layers. In transformer architectures, cross-attention provides the most natural and expressive mechanism for ``sending features'' between distant layers, offering superior semantic understanding and selective information transfer compared to naive concatenation or addition operations.

Unlike concatenation, which simply combines feature representations without considering their relative importance or relevance, cross-attention allows the decoder to selectively attend to relevant encoder features based on the current decoding context. This selective attention mechanism is particularly beneficial for text-to-image generation, where different spatial regions may require different types and amounts of information from the encoder path.

\subsection{Text Encoder Strategy and Architectural Minimalism}

\textbf{Text Encoding Approach.} Text encoding represents a critical component in text-to-image generation, significantly influencing both generation quality and computational efficiency. While early models relied on CLIP encoders ~\citep{rombach2021highresolution, podell2024sdxl} and recent models incorporate T5-series models ~\citep{li2024hunyuanditpowerfulmultiresolutiondiffusion, esser2024scalingrectifiedflowtransformers, flux2024}, we observe an emerging trend toward using causal language models as text encoders. ~\citep{qin2025luminaimage20unifiedefficient, wu2025qwenimagetechnicalreport}

Recent research demonstrates that fine-tuned large language models can effectively serve as embedding models. For instance, Qwen has released dedicated embedding models based on their language model architectures, showing promising results in various downstream tasks.

In HDM, we employ Qwen3-0.6B ~\citep{yang2025qwen3technicalreport} as our text encoder, representing one of the smallest causal language models suitable for this task while maintaining reasonable text understanding capabilities. We preserve the unidirectional attention mechanism and directly utilize the final hidden state as our text embedding, leveraging the implicit positional information inherent in causal attention patterns.

This choice balances computational efficiency with text understanding capability, enabling effective text-image alignment without the computational overhead of larger text encoders.

\textbf{Minimal Architecture Design.} Modern text-to-image models employ various sophisticated strategies for handling text and image modalities separately. DiT uses cross-attention to inject text features into image features, while MMDiT employs separate adaptive layer normalization (adaLN) modulation and MLPs with joint attention mechanisms ~\citep{esser2024scalingrectifiedflowtransformers}.

HDM pursues architectural minimalism by directly concatenating text and image features as input to the entire backbone network. This approach aligns with UViT's philosophy of incorporating class tokens and timestep tokens into the input sequence, though we retain adaLN for conditional information to maintain generation quality.

\textbf{Shared Adaptive Layer Normalization.} Following recent efficiency-focused approaches like DiT-Air ~\citep{chen2025ditairrevisitingefficiencydiffusion} and Chroma, we implement shared adaLN across all layers to reduce parameter count while preserving conditioning effectiveness.

Traditional adaLN applies layer-specific modulation:
\begin{align}
\text{adaLN}^{(l)}(x, c) = \gamma^{(l)}(c) \odot \frac{x - \mu(x)}{\sigma(x)} + \beta^{(l)}(c)
\end{align}
where $\gamma^{(l)}(c)$ and $\beta^{(l)}(c)$ are layer-specific learned functions of condition $c$.

Our shared adaLN uses global parameters across all layers:
\begin{align}
\text{adaLN}_{\text{shared}}(x, c) = \gamma(c) \odot \frac{x - \mu(x)}{\sigma(x)} + \beta(c)
\end{align}
where $\gamma(c)$ and $\beta(c)$ are generated by a single learned MLP from input conditions (timestep and other conditioning information). This strategy significantly reduces parameters while preserving performance ~\citep{chen2025ditairrevisitingefficiencydiffusion}.

\subsection{Positional Encoding Strategy}

Effective positional encoding is crucial for enabling arbitrary aspect ratio generation while maintaining spatial coherence. For spatial position encoding in $H \times W$ images, we establish specific mathematical constraints on the range parameters to ensure consistent behavior across different aspect ratios.

For H-axis range ($r_H$) and W-axis range ($r_W$), we enforce:
\begin{align}
r_H \times r_W &= 1.0 \label{eq:constraint1}\\
\frac{r_H}{r_W} &= \frac{H}{W} \label{eq:constraint2}
\end{align}

Solving these constraints simultaneously yields:
\begin{align}
r_H = \sqrt{\frac{H}{W}}, \quad r_W = \sqrt{\frac{W}{H}}
\end{align}

This formulation enables arbitrary aspect ratio handling while maintaining consistent positional encoding properties across different image dimensions. We apply 2D axial Rotary Position Embedding (RoPE) ~\citep{10.1016/j.neucom.2023.127063} to the first half of the feature dimensions:
\begin{align}
\text{RoPE}(x, \text{pos}) = \begin{pmatrix} 
x_1 \cos(\text{pos}/\theta_1) - x_2 \sin(\text{pos}/\theta_1) \\ 
x_1 \sin(\text{pos}/\theta_1) + x_2 \cos(\text{pos}/\theta_1) \\ 
\vdots 
\end{pmatrix}
\end{align}

For text sequences, we employ NoPE (No Position Embedding) ~\citep{wang2024lengthgeneralizationcausaltransformers}, as the causal attention mechanism's inherent property of later tokens attending to previous tokens provides sufficient implicit positional information for our text understanding needs.

\subsection{Shifted Square Crop Training Strategy}

\begin{figure*}[h]
\centering
\includegraphics[width=0.8\textwidth]{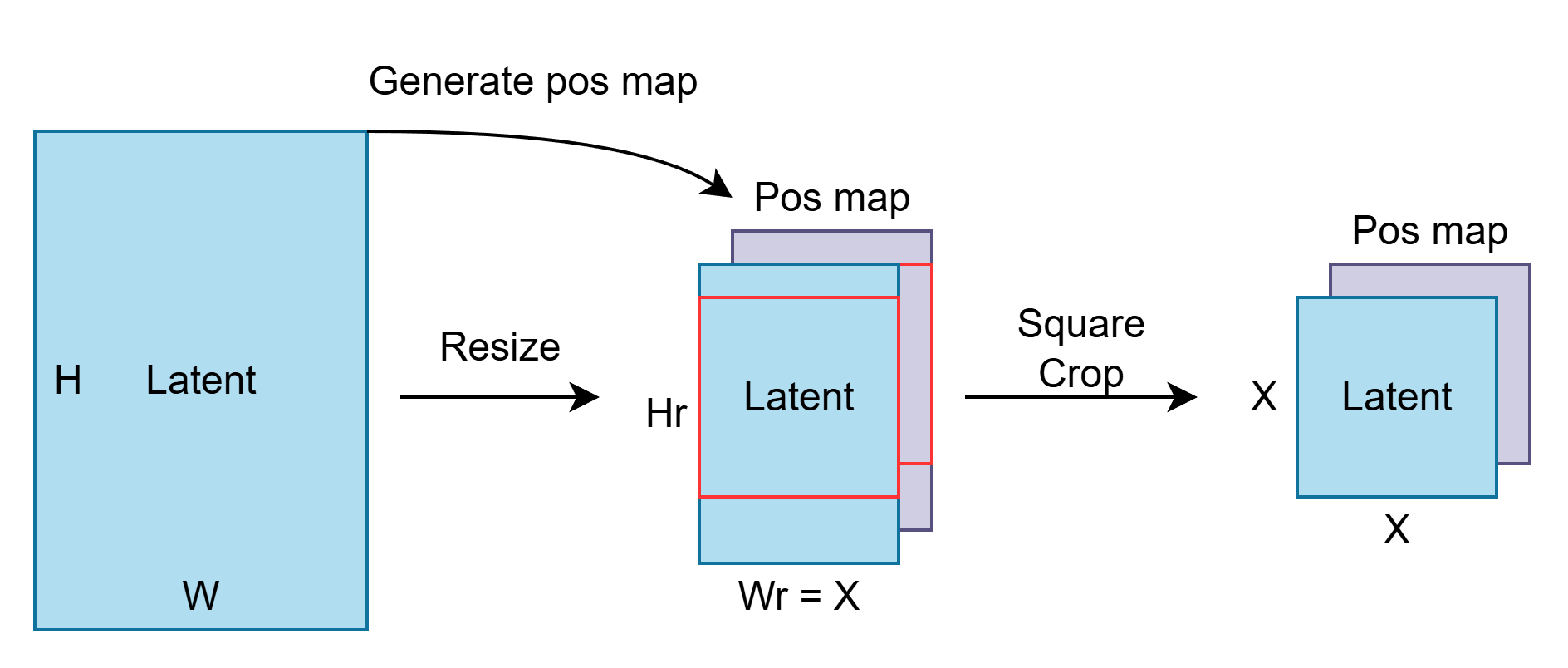}
\caption{An visualized example of position map creation and shifted square crop workflow.}
\label{fig:sq-crop}
\end{figure*}

Aspect ratio handling during training presents significant computational and implementation challenges, particularly for text-to-image tasks where users expect generation at arbitrary aspect ratios. Standard Aspect Ratio Bucketing (ARB) approaches group images with similar aspect ratios into batches, but require complex batch pre-computation, specialized dataset implementations, and careful memory management.

We propose \textbf{Shifted Square Crop}, inspired by patch diffusion pretraining techniques ~\citep{wang2023patch}. This approach enables efficient arbitrary aspect ratio training while maintaining computational simplicity and memory efficiency. As showing in \ref{fig:sq-crop}, we apply following strategy:

\textbf{Training Procedure.} Given an original image $I_{\text{orig}}$ with dimensions $(H_{\text{orig}}, W_{\text{orig}})$ and target training size $X$, our procedure follows these steps:

\textbf{Step 1 - Resize:} Scale the image such that $\min(H_{\text{orig}}, W_{\text{orig}}) = X$:
\begin{align}
I_{\text{resized}} = \text{Resize}\left(I_{\text{orig}}, \text{scale}=\frac{X}{\min(H_{\text{orig}}, W_{\text{orig}})}\right)
\end{align}

\textbf{Step 2 - Position Mapping:} Create position map $P \in \mathbb{R}^{H_{\text{resized}} \times W_{\text{resized}} \times 2}$ where:
\begin{align}
P[i,j] = \left(\frac{i \cdot r_H}{H_{\text{resized}}}, \frac{j \cdot r_W}{W_{\text{resized}}}\right)
\end{align}

\textbf{Step 3 - Random Cropping:} Select random crop coordinates $(x_0, y_0)$ and extract:
\begin{align}
I_{\text{crop}} &= I_{\text{resized}}[y_0:y_0+X, x_0:x_0+X]\\
P_{\text{crop}} &= P[y_0:y_0+X, x_0:x_0+X]
\end{align}

\textbf{Step 4 - Training:} Use the $(I_{\text{crop}}, P_{\text{crop}})$ pair for diffusion training.

During inference, we utilize the uncropped position map $P_{\text{full}}$ corresponding to the target generation aspect ratio, enabling arbitrary aspect ratio generation following principles established in patch diffusion methodologies.

\subsection{TREAD Integration for Acceleration}

To accelerate convergence and improve step efficiency under resource constraints, we integrate TREAD (Token Routing for Efficient Architecture-agnostic Diffusion) ~\citep{krause2025treadtokenroutingefficient}. While alternative approaches like REPA ~\citep{yu2025representation} or REPA-E ~\citep{leng2025repaeunlockingvaeendtoend} offer convergence improvements, they require additional ViT models for reference feature generation and extra MLPs for feature fitting, significantly reducing overall computational efficiency. REPA-E particularly demands additional VAE backward passes, further increasing computational overhead.

TREAD provides superior efficiency by implementing architecture-agnostic token routing without requiring auxiliary models or external dependencies, making it ideally suited for resource-constrained training scenarios. The method selectively processes only a fraction of tokens during training while maintaining gradient flow to all parameters, achieving both memory efficiency and convergence acceleration.

\subsection{Dataset and Preprocessing}

For our initial HDM validation, we utilize Danbooru2023, a high-quality anime-style image-tag dataset containing approximately 7.6M images. This dataset choice reflects our focus on demonstrating consumer-level training feasibility while providing a challenging and well-defined generation domain with rich semantic diversity.

We employ Pixtral-11M, a state-of-the-art vision-language model, to generate natural language captions for each image in the dataset. These generated captions serve as text conditions during training, replacing or supplementing the original tag-based annotations. This preprocessing step significantly enhances the model's ability to understand and generate images based on natural language descriptions, improving user experience and prompt adherence.

The choice of natural language captions over structured tags represents a design decision favoring accessibility and user-friendliness, enabling more intuitive interaction patterns for end users while maintaining the semantic richness necessary for high-quality generation.

\subsection{Inference Methodology and Camera Control}

Our training methodology necessitates specific inference procedures that unlock unique generation capabilities not commonly found in other text-to-image models.

\textbf{Setup Requirements:}
\begin{itemize}
\item Disable TREAD selection during inference to ensure full model capacity utilization
\item Generate random noise $X_1$ matching the target image dimensions
\item Create full-size position map corresponding to desired aspect ratio
\item Perform flow matching sampling with appropriate step scheduling
\end{itemize}

\textbf{Position Map Manipulation for Camera Control.} Due to training with cropped position maps, our model exhibits interesting emergent behaviors enabling intuitive ``camera'' control during generation. The Danbooru dataset's aspect ratio distribution bias means horizontal image generation may naturally produce compositions that appear as crops from larger vertical or square images.

However, position map manipulation enables precise compositional control:

\begin{itemize}
\item \textbf{X-axis Translation:} Positive values shift the ``camera'' rightward, negative values shift leftward
\item \textbf{Y-axis Translation:} Positive values move the ``camera'' downward, negative values move upward  
\item \textbf{Zoom Control:} Values greater than 1.0 zoom in (crop effect), values less than 1.0 zoom out (wider view)
\end{itemize}

This capability emerges naturally from our shifted square crop training strategy and sophisticated positional encoding approach, providing users with fine-grained compositional control without requiring additional training or architectural modifications.

\textbf{Auto-Guidance via TREAD Selection.} Intuitively, a model with TREAD selection enabled can be viewed as a deliberately weakened version of the full model. Following the Auto-Guidance paradigm ~\citep{karras2024guiding, chen2025s2guidancestochasticselfguidance} and recommendations from the official TREAD implementation ~\citep{tread_inference_code}, we employ differential TREAD selection rates for conditional (cr) and unconditional (ur) guidance such that $\text{cr} < \text{ur}$ and $\text{cr} < 0.5$ (the TREAD selection rate used during training).

This approach provides improved guidance without requiring separate model training or additional computational overhead during inference.

%% file: sections/training.tex
\section{Training}
\label{sec:training}

This section details our comprehensive training approach, from model scaling decisions to cost analysis, demonstrating the feasibility of high-quality text-to-image model development on consumer-grade hardware.

\subsection{Model Architecture Specifications}

\textbf{Latent Encoder/Decoder (VAE) Choice}
For latent diffusion model, the choice of VAE is critical, particularly for a model with a constrained parameter budget like HDM. While the standard VAE offers high-fidelity reconstruction, their latent space presents two primary challenges for our use case.

First, recent work suggests that models with limited capacity can struggle when paired with VAEs that have a large latent channel dimension, potentially leading to degraded performance~\citep{esser2024scalingrectifiedflowtransformers}. Second, the latent distributions of standard VAEs can be complex or "noisy," making the subsequent denoising task more difficult for the diffusion model. A smoother latent space with properties more aligned with natural images is considered more "diffusable" and thus easier to model~\citep{kouzelis2025eqvaeequivarianceregularizedlatent, skorokhodov2025improvingdiffusabilityautoencoders}.

To mitigate these issues, we use the VAE of SDXL model~\citep{rombach2021highresolution, podell2024sdxl} and employ the Entropy-Quantized VAE (EQ-VAE) methodology to fine-tune the base SDXL VAE~\citep{kouzelis2025eqvaeequivarianceregularizedlatent}. This approach regularizes the VAE to produce a smoother latent space whose frequency distribution more closely resembles that of natural images. By doing so, we create a more favorable starting point for our diffusion backbone, simplifying the generative task and allowing the XUT model to allocate its capacity more efficiently.

\textbf{XUT variants}

To achieve efficient pretraining and inference on consumer hardware while maintaining competitive generation quality, we carefully balance model size with capability requirements. We propose three HDM backbone scales to accommodate different computational budgets and performance targets.

\begin{table*}[h]
\centering
\begin{tabular}{@{}lccc@{}}
\toprule
 & \textbf{XUT-small} & \textbf{XUT-base} & \textbf{XUT-large} \\
\midrule
Dimension & 896 & 1024 & 1152  \\
Context Dim (Text Encoder) & 640 (Gemma3-270M) & 1024 (Qwen3-0.6B) & 1152 (Gemma3-1B) \\
MLP Dimension & 3072 & 3072 & 5120 \\
Attention Heads & 14 & 16 & 12 \\
Attention Head Dimension & 64 & 64 & 96 \\
XUT Depth ($n_{\text{depth}}$) & 4 & 4 & 4 \\
Encoder Blocks ($n_{\text{enc}}$) & 1 & 1 & 1  \\
Decoder Blocks ($n_{\text{dec}}$) & 2 & 3 & 3\\
TREAD Depth Before ($N$) & 1 & 1 & 1\\
TREAD Depth After ($M$) & 3 & 3 & 3  \\
\midrule
Total Transformer Blocks & 16 & 20 & 20 \\
Total Attention Layers & 20 & 24 & 24 \\
Sequence Length at 256² & 256 & 256 & 256  \\
Parameters (XUT only) & 230M & 343M & 550M \\
\bottomrule
\end{tabular}
\caption{HDM model architecture specifications across three scales. XUT-base serves as our primary validation target, balancing efficiency with generation quality.}
\label{tab:model_specs}
\end{table*}

The architecture specifications ~\ref{tab:model_specs} reflect careful optimization for the consumer hardware training constraint. XUT-base represents our primary validation target, providing an optimal balance between model capacity and computational efficiency. This technical report focuses on XUT-base validation to establish complete training recipes and validate our design principles.

\subsection{Progressive Training Recipe}

Our XUT-base training employs a progressive resolution scaling strategy, beginning at 256² resolution and advancing through multiple stages to achieve high-resolution generation capability. We utilize four RTX5090 GPUs with distributed data parallel (DDP) training, demonstrating scalability within consumer hardware constraints.

\begin{table*}[h]
\centering

\begin{threeparttable}
\begin{tabular}{@{}lcccc@{}}
\toprule
\textbf{Stage} & \textbf{256²} & \textbf{512²} & \textbf{768²} & \textbf{1024²} \\
\midrule
\midrule
\multicolumn{5}{c}{\textit{Dataset Configuration}} \\
\midrule
Dataset & Danbooru2023 & \multicolumn{2}{c}{ Danbooru2023 + Extra\textsuperscript{*}} & Curated\textsuperscript{**} \\
Image Count & 7.624M & 8.469M & 8.469M & 3.649M \\
Epochs & 20 & 5 & 1 & 1  \\
Samples Seen & 152.5M & 42.34M & 8.469M & 3.649M \\
Patches Seen & 39.0B & 43.5B & 19.5B & 15.0B \\
\midrule
\midrule
\multicolumn{5}{c}{\textit{Training Configuration}} \\
\midrule
Learning Rate (μP, base\_dim=1) & 0.5 & 0.1 & 0.05 & 0.02  \\
Batch Size (per GPU) & 128 & 64 & 64 & 16  \\
Gradient Checkpointing & No & Yes & Yes & Yes  \\
Gradient Accumulation & 4 & 2 & 2 & 4  \\
Global Batch Size & 2048 & 512 & 512 & 256  \\
TREAD Selection Rate & 0.5 & 0.5 & 0.5 & 0.0  \\
Context Length & 256 & 256 & 256 & 512  \\
\midrule
\midrule
\multicolumn{5}{c}{\textit{Resource Utilization}} \\
\midrule
Training Wall Time (hour) & 174 & 120 & 42 & 49  \\
\bottomrule
\end{tabular}
\caption{Comprehensive training recipe for XUT-base model across progressive resolution stages.}
\label{tab:training_recipe}
\footnotesize
\begin{tablenotes}
    \item[*] Extra dataset includes internal collections such as PVC figure photographs and filtered Pixiv artist sets.
    \item[**] Curated dataset uses Danbooru2023 filtered for quality indicators: ``masterpiece,'' ``best quality'' vs. ``low quality,'' ``worst quality,'' and temporal tags ``newest,'' ``recent'' vs. ``old'' to enhance negative prompting effectiveness.
\end{tablenotes}
\end{threeparttable}
\end{table*}

\textbf{Progressive Resolution Strategy.} Our progressive training approach addresses the computational challenges of high-resolution training while maintaining generation quality. Early stages focus on learning fundamental visual concepts and text-image alignment at lower resolutions, while later stages refine high-frequency details and spatial coherence.

The learning rate schedule follows μP (Maximal Update Parameterization) principles with base dimension normalization, ensuring stable training dynamics across different model scales and resolutions. TREAD selection rates are maintained at 0.5 throughout most training stages, with full model capacity utilized only in the final 1024² stage following author recommendations for improved CFG generation quality.

\textbf{Dataset Evolution.} Our dataset strategy evolves across training stages to address specific requirements at each resolution level. Early stages utilize the full Danbooru2023 dataset to learn broad visual concepts, while later stages employ curated subsets optimized for high-quality generation and effective negative prompting.

The curated 1024² dataset specifically filters for quality indicators and temporal relevance, enhancing the model's ability to generate high-quality images while responding appropriately to negative prompts—a crucial capability for practical deployment.

\subsection{Training Cost Analysis and Accessibility}

One of our primary contributions is demonstrating that high-quality text-to-image models can be trained at dramatically reduced costs compared to traditional data center approaches. Our comprehensive cost analysis provides a roadmap for researchers with limited computational resources.

\textbf{Hardware Cost Analysis.} Based on vast.ai pricing (as of 2025), renting four RTX5090 rigs for the complete 385-hour training duration costs approximately \$535-620, including regional pricing variations and availability fluctuations. This represents a dramatic reduction compared to previous data center GPU costs of \$1000-1500 for comparable training runs .

Our pretraining utilizes self-hosted consumer-level hardware, establishing a new benchmark for cost efficiency in 1024×1024 text-to-image model training. The total hardware investment for a complete training setup (four RTX5090 GPUs with supporting infrastructure) approximates \$8000-10000, enabling multiple training runs and iterative development.

\textbf{Storage and Memory Considerations.} An important practical consideration involves storage requirements for efficient training. All training proceeds without latent caching or text embedding caching, as full dataset caching would require storage exceeding typical consumer setups.

With f16/bf16 precision, caching all text embeddings and f8c16 latents for our Danbooru dataset would require over 12TB of high-speed storage, while the original dataset occupies only 1.4-1.8TB. This storage requirement would significantly increase infrastructure costs and complexity, potentially negating the accessibility advantages of our approach.

Future plans involving larger datasets (40M+ images) would further exacerbate storage requirements, making on-the-fly processing a necessity for maintaining cost effectiveness and accessibility.

\textbf{Energy Efficiency.} Our training approach achieves competitive generation quality while consuming significantly less energy than large-scale data center training runs. The complete training cycle consumes approximately 125 kWh across four RTX5090 GPUs, equivalent to standard household energy consumption for 4-5 days, demonstrating environmental sustainability alongside cost efficiency.

%% file: sections/result.tex
\section{Results}
\label{sec:results}

\subsection{Prompt Following with Composition Consistency}

We observed that HDM achieves a remarkable capability: \textit{composition consistency under prompt modifications}. Specifically, the model maintains the global composition when tags are changed without adding or removing concepts, under the same seed and sampler settings. This property enables fine-grained control over specific attributes while preserving the overall image structure.

\begin{table}[h]
\centering
\resizebox{\textwidth}{!}{
\begin{tabular}{lcccc}
\toprule
\textbf{Prompt} & \textbf{Base} & \textbf{+ neg: ``jacket''} & \textbf{+ neg: ``open jacket''} & \textbf{``open'' $\rightarrow$ ``closed mouth''} \\
\midrule
\textbf{Image} & \includegraphics[width=0.18\textwidth,trim={0 800pt 0 0},clip]{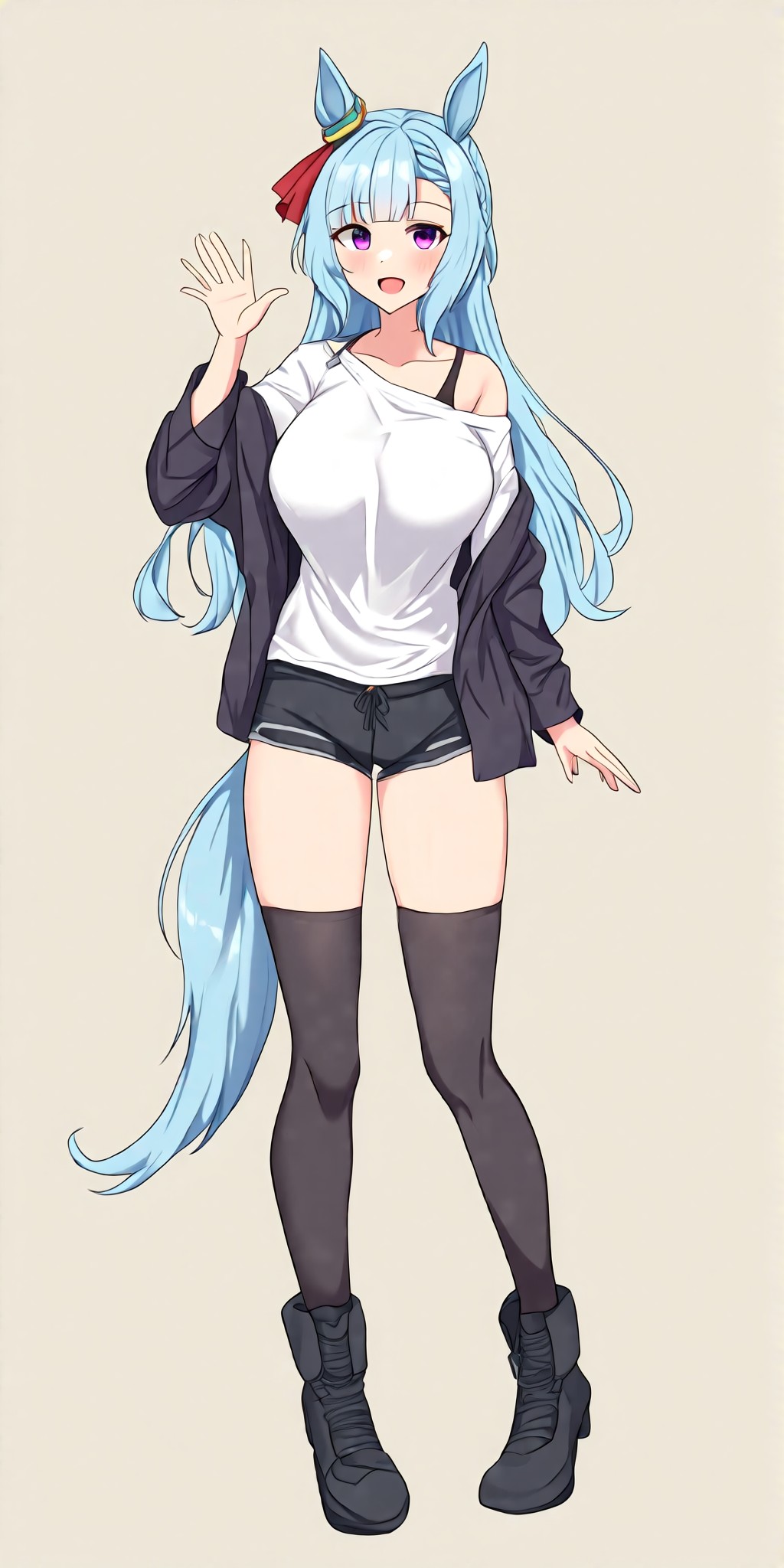} & 
\includegraphics[width=0.18\textwidth,trim={0 800pt 0 0},clip]{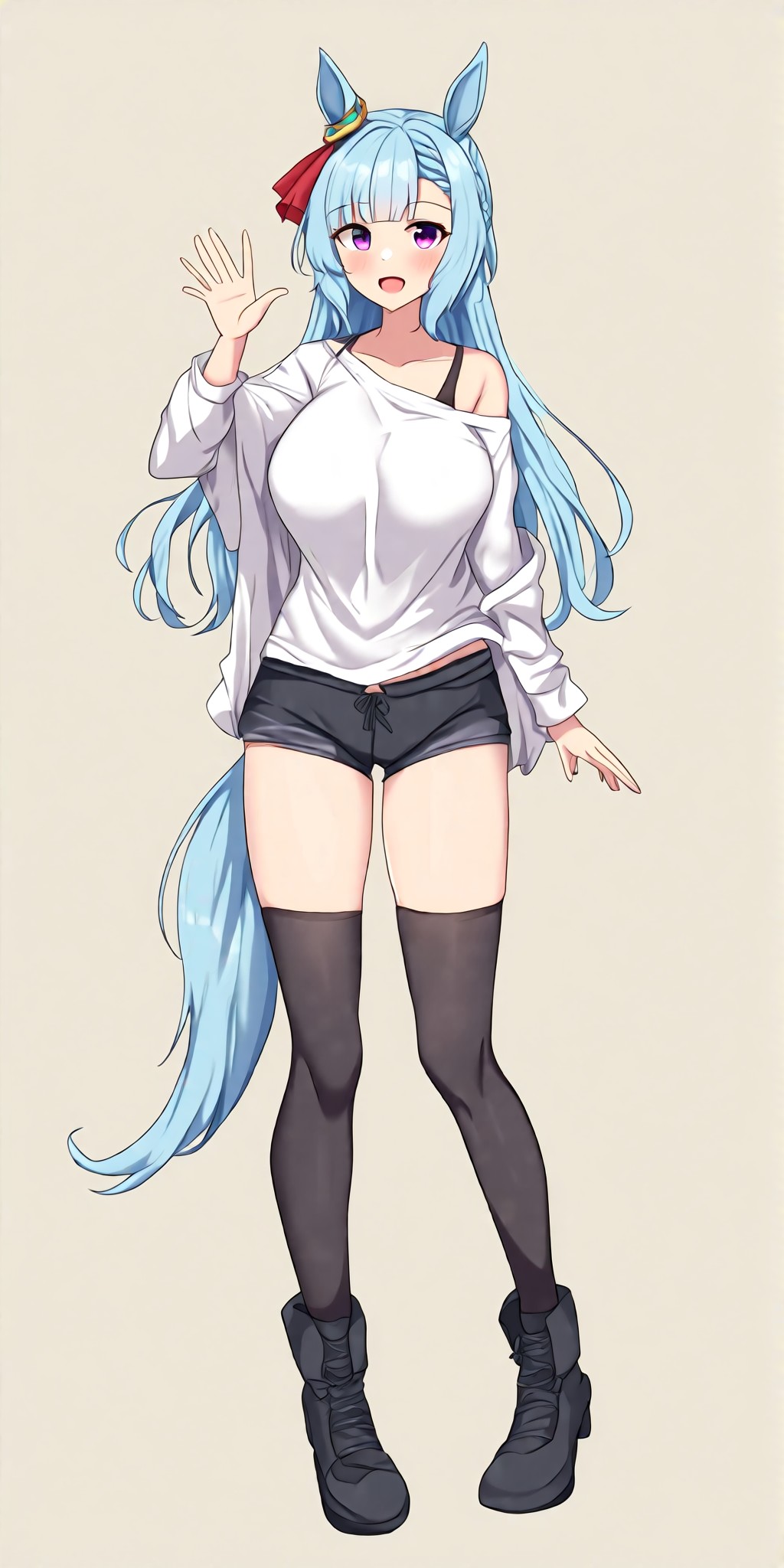} & 
\includegraphics[width=0.18\textwidth,trim={0 800pt 0 0},clip]{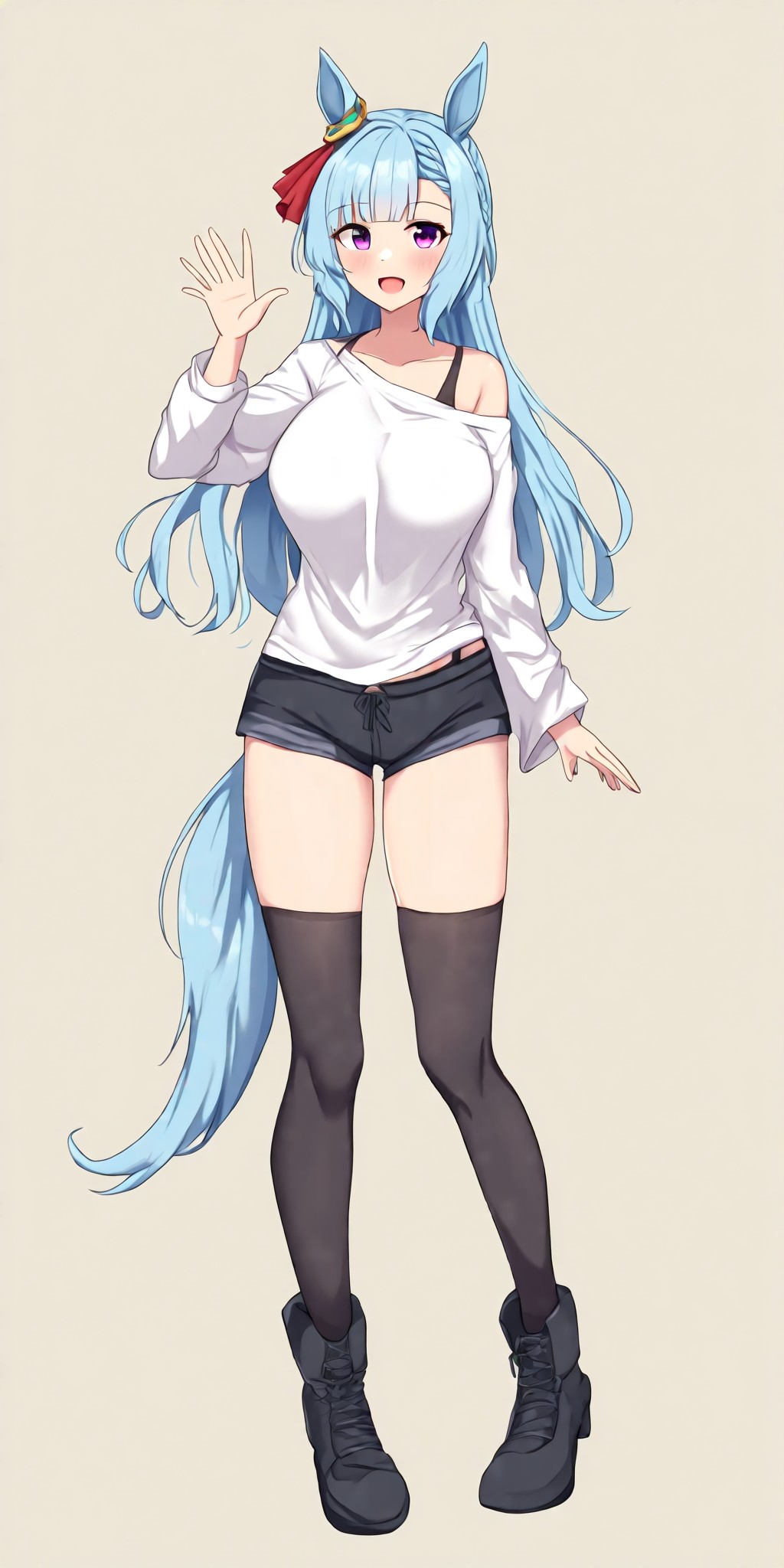} &
\includegraphics[width=0.18\textwidth,trim={0 800pt 0 0},clip]{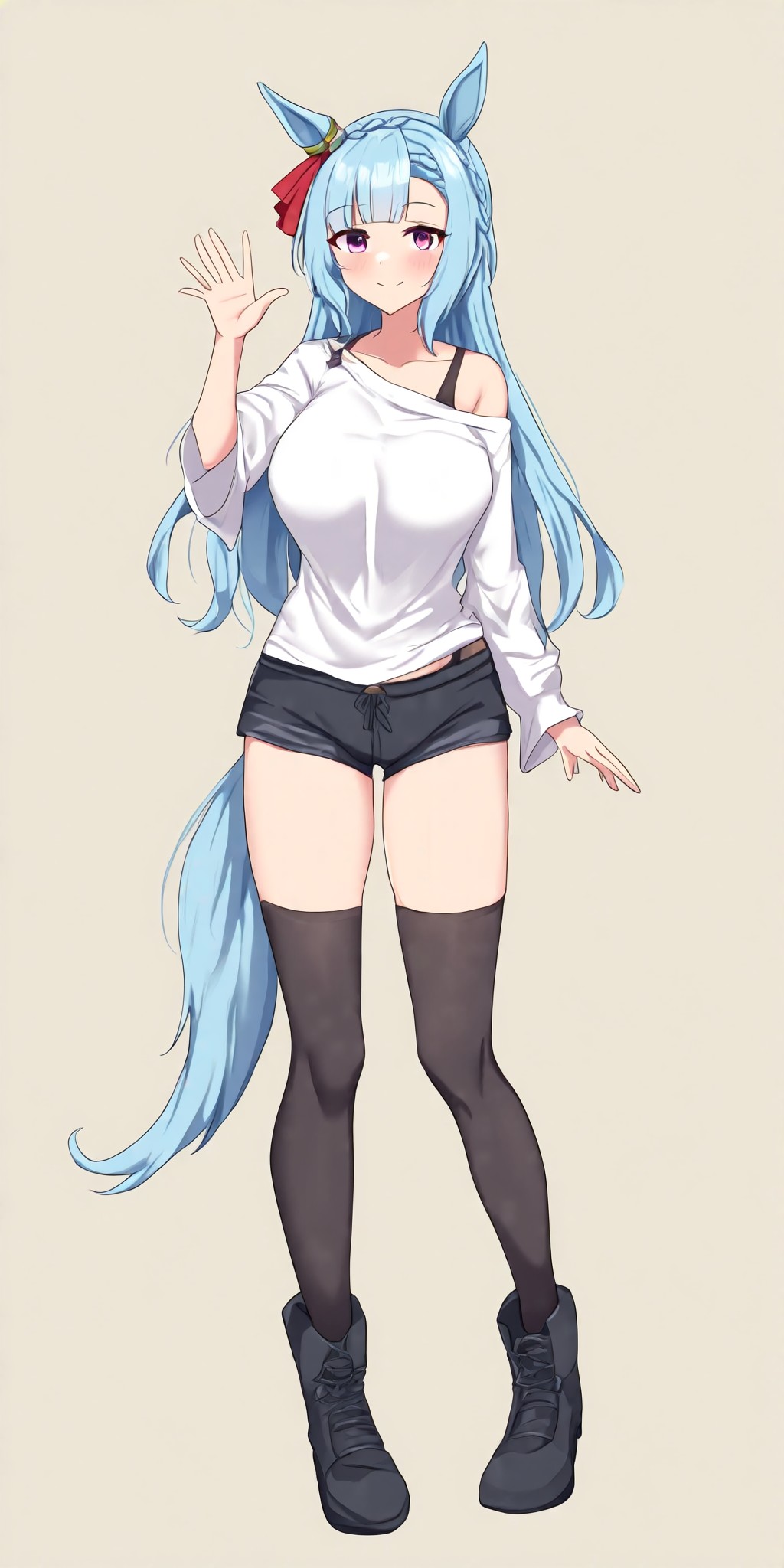} \\
\bottomrule
\end{tabular}
}
\caption{Composition consistency results1.}
\label{tab:composition_detailed}
\end{table}

\begin{table}[!htbp]
\centering
\resizebox{\textwidth}{!}{
\begin{tabular}{lccc}
\toprule
\textbf{Prompt} & \textbf{Base} & \textbf{``closed'' $\rightarrow$ ``open mouth''} & \textbf{``white'' $\rightarrow$ ``black footwear''} \\
\midrule
\textbf{Image} & \includegraphics[width=0.25\textwidth,trim={0 0pt 0 0},clip]{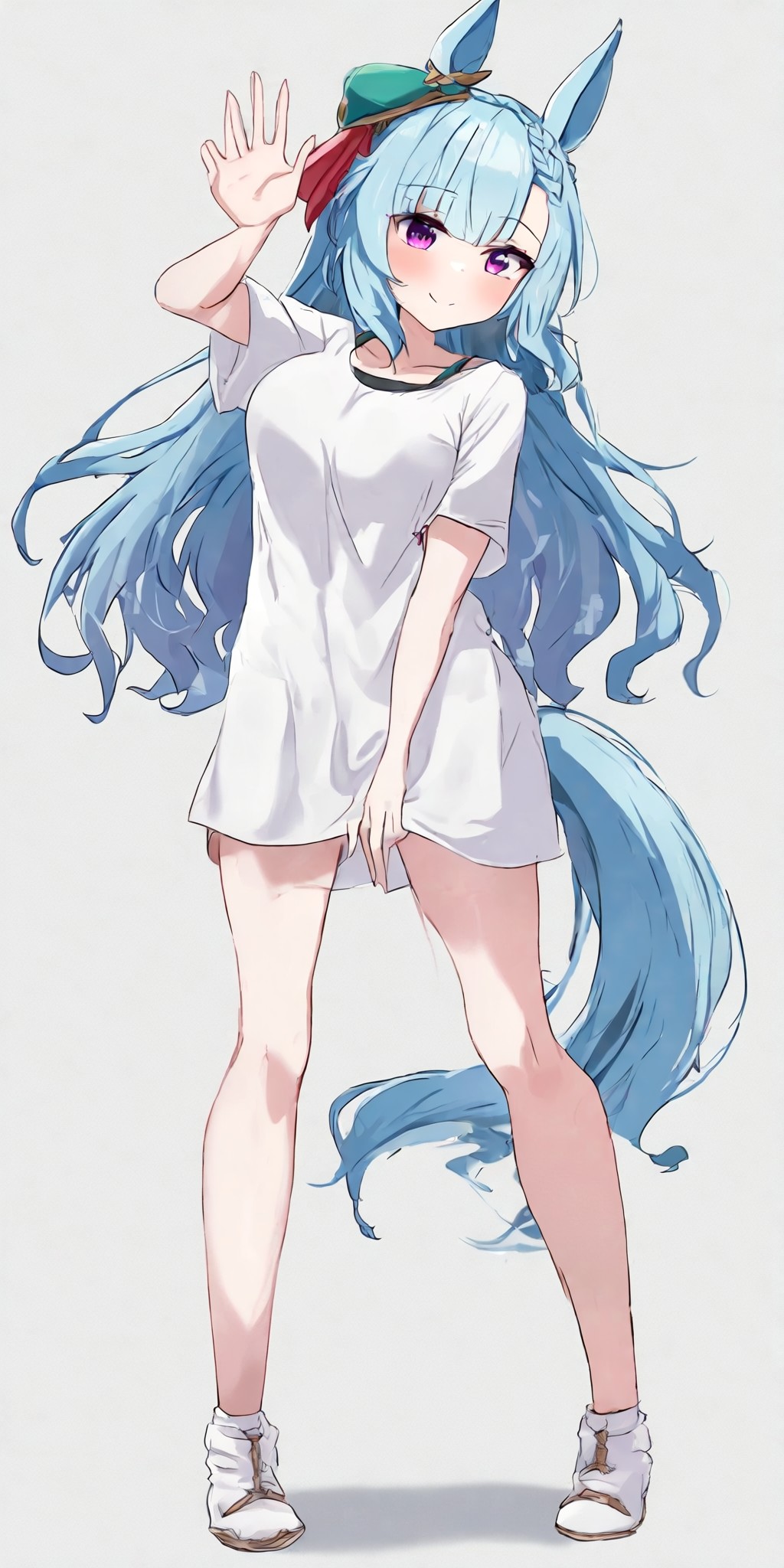} & 
\includegraphics[width=0.25\textwidth,trim={0 0pt 0 0},clip]{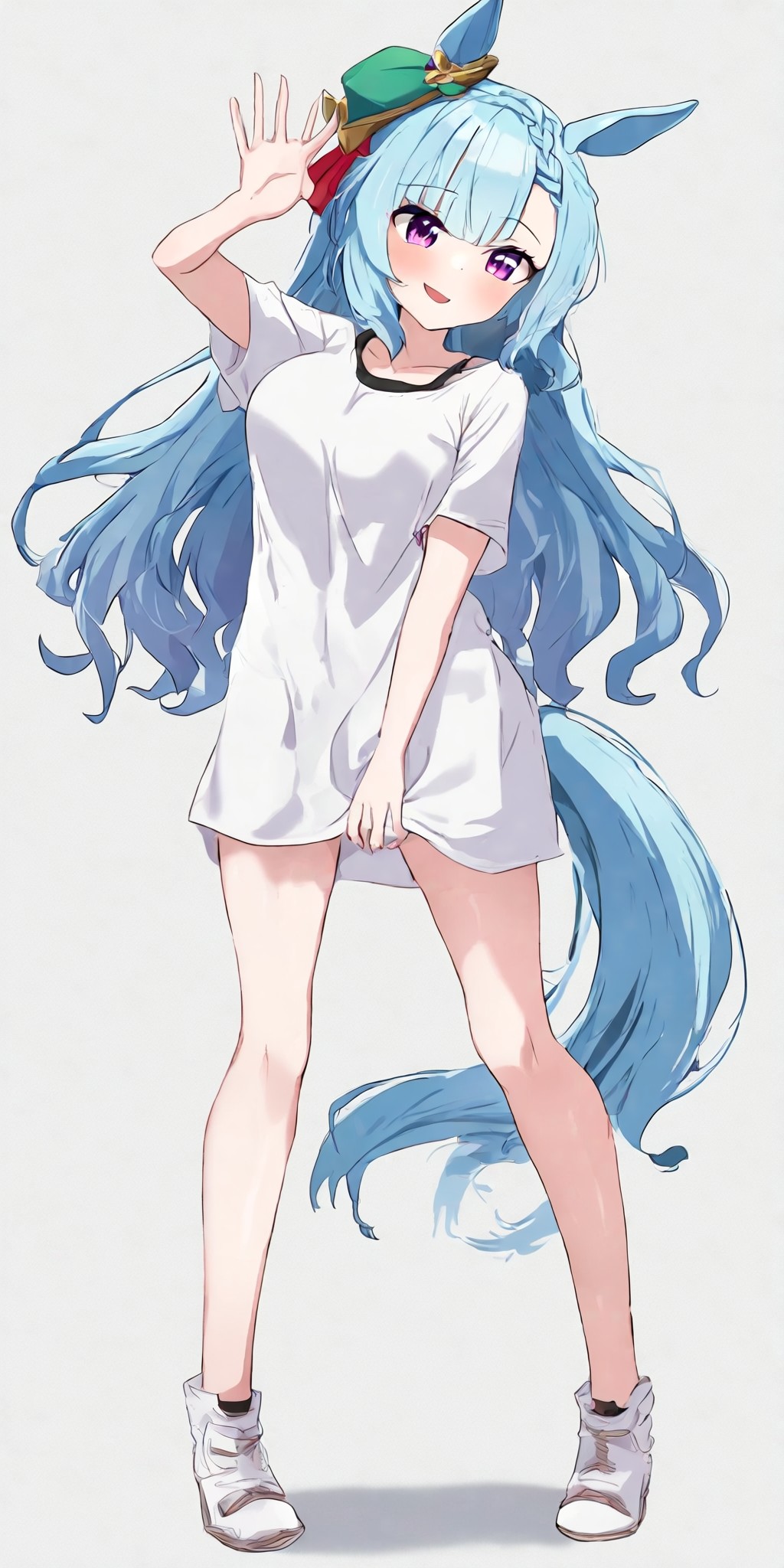} & 
\includegraphics[width=0.25\textwidth,trim={0 0pt 0 0},clip]{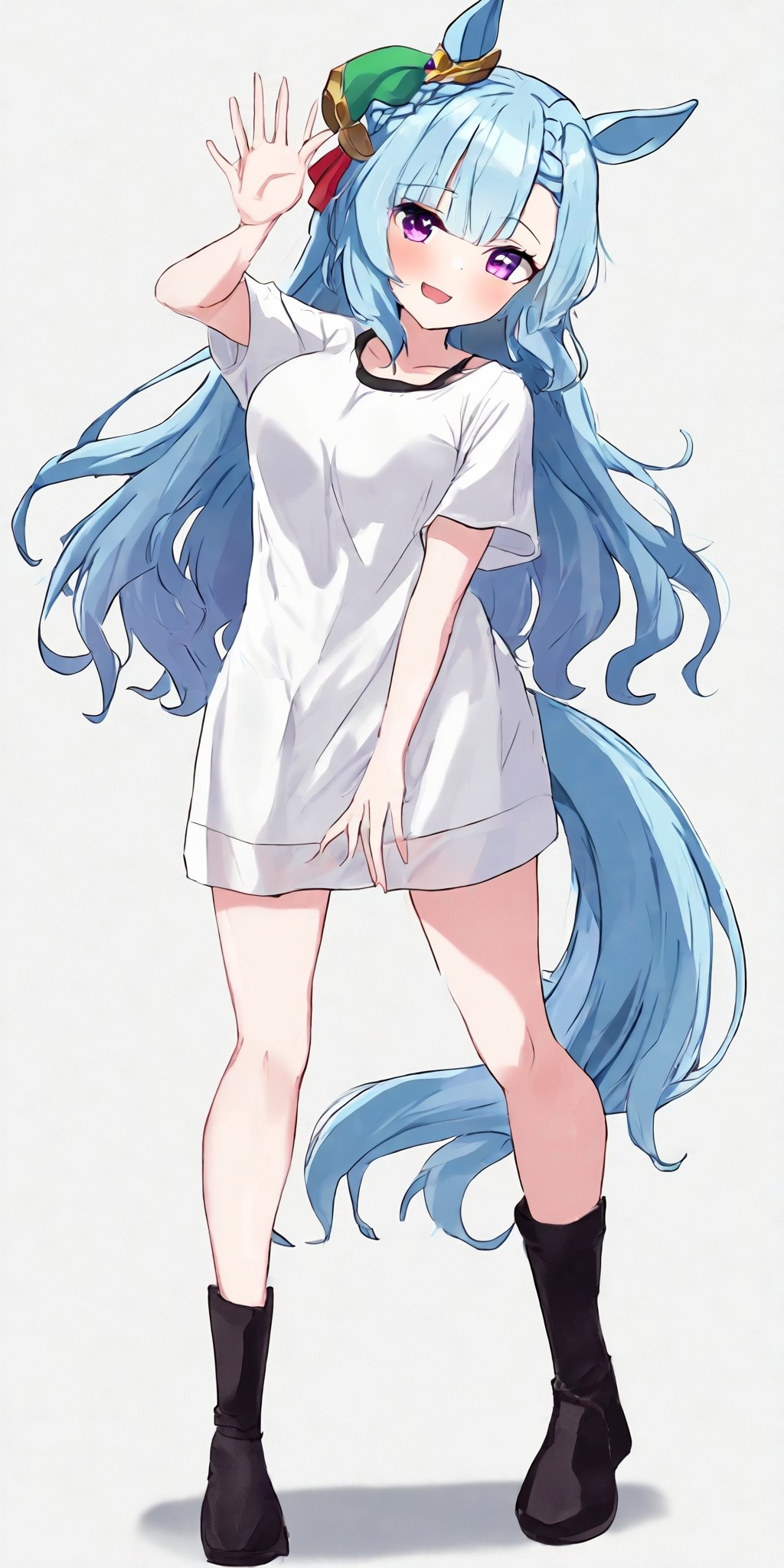} \\
\bottomrule
\end{tabular}
}
\caption{Composition consistency results2.}
\label{tab:composition_simple}
\end{table}

As shown in \ref{tab:composition_detailed} and \ref{tab:composition_simple}, we can find our model provide surprisingly good composition consistency when the modified part of prompt is not related to overall subjects or topic.

\subsection{Position Map Manipulation Results}

Our shifted square crop training strategy enables intuitive camera control through position map manipulation. The following results demonstrate the effectiveness of this approach across three control dimensions: horizontal shift, vertical shift, and zoom.

\begin{table}[h]
\centering
\begin{tabular}{lccc}
\toprule
\textbf{X-Shift} & \textbf{+0.25 (Right)} & \textbf{No Shift} & \textbf{-0.25 (Left)} \\
\midrule
\textbf{Example 1} & \includegraphics[width=0.25\textwidth,trim={0 500pt 0 100pt},clip]{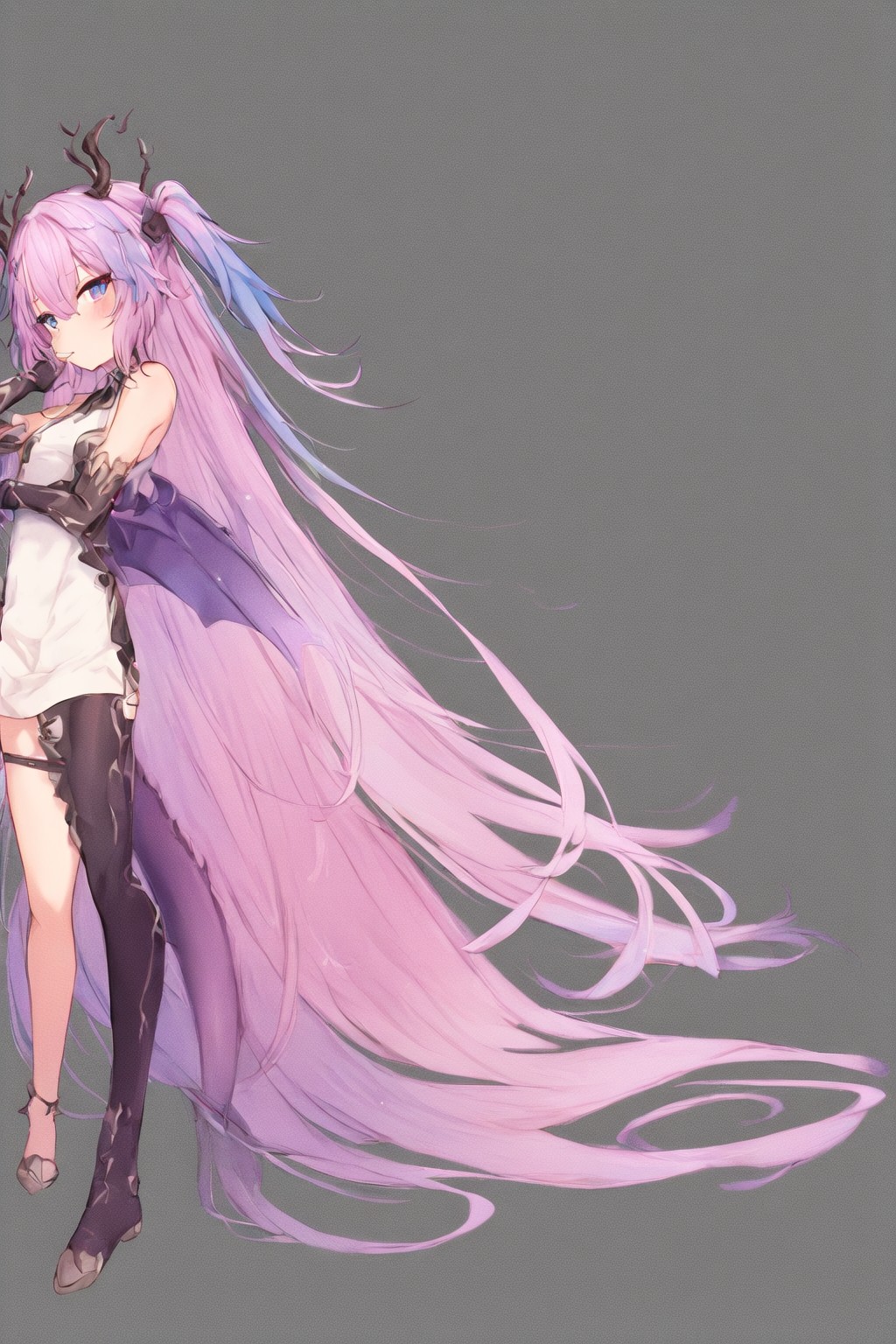} & 
\includegraphics[width=0.25\textwidth,trim={0 500pt 0 100pt},clip]{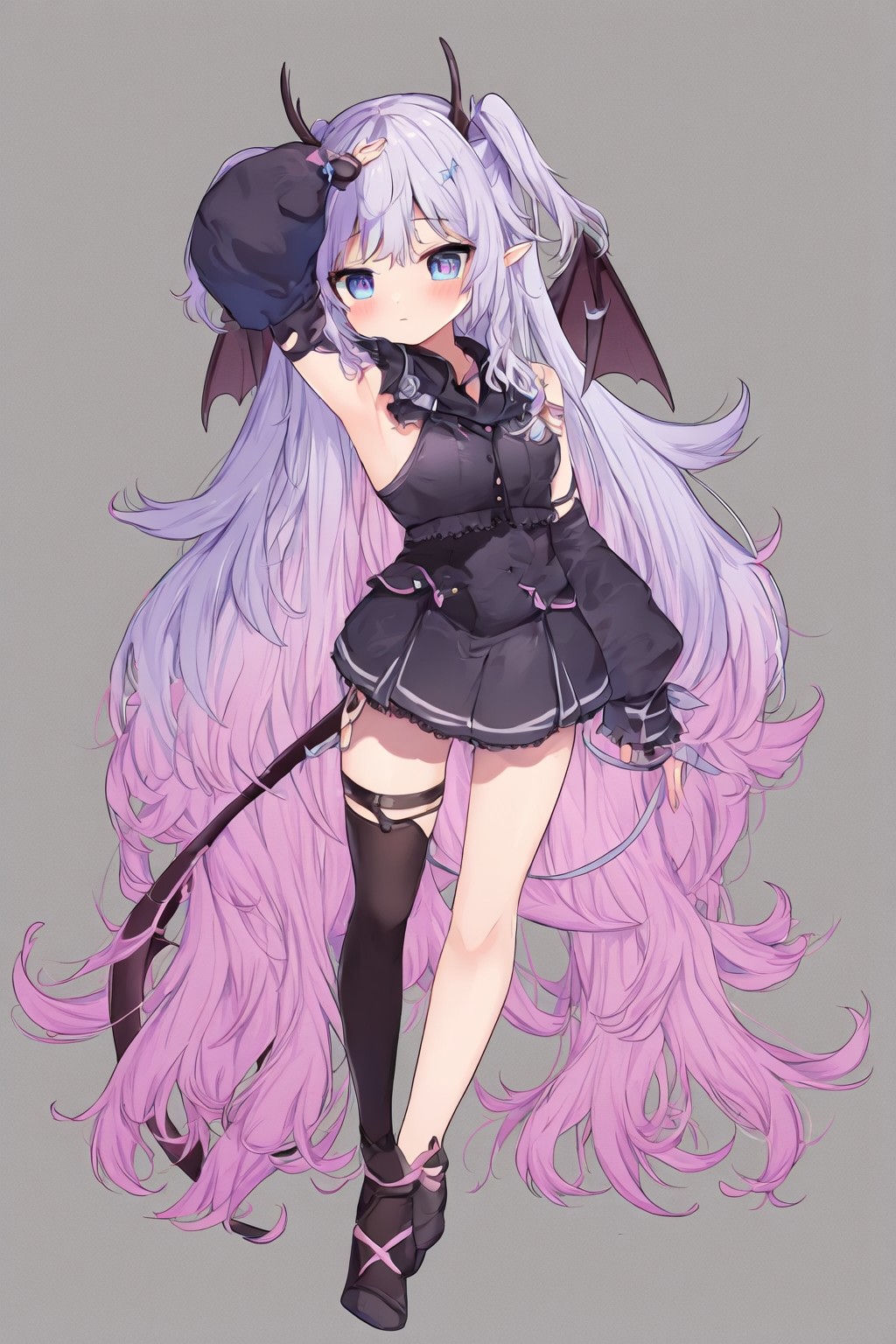} & 
\includegraphics[width=0.25\textwidth,trim={0 500pt 0 100pt},clip]{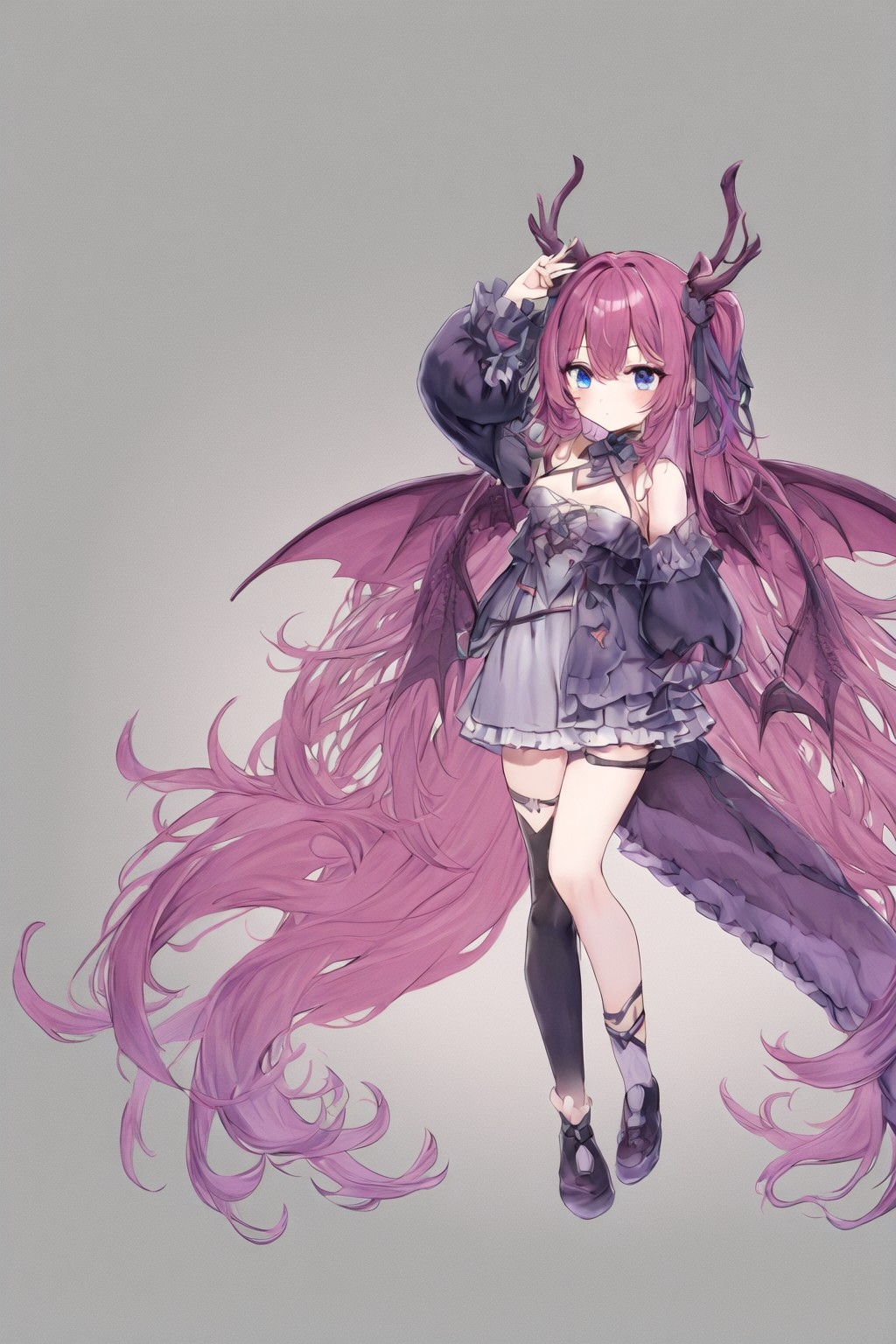} \\
\textbf{Example 2} & \includegraphics[width=0.25\textwidth]{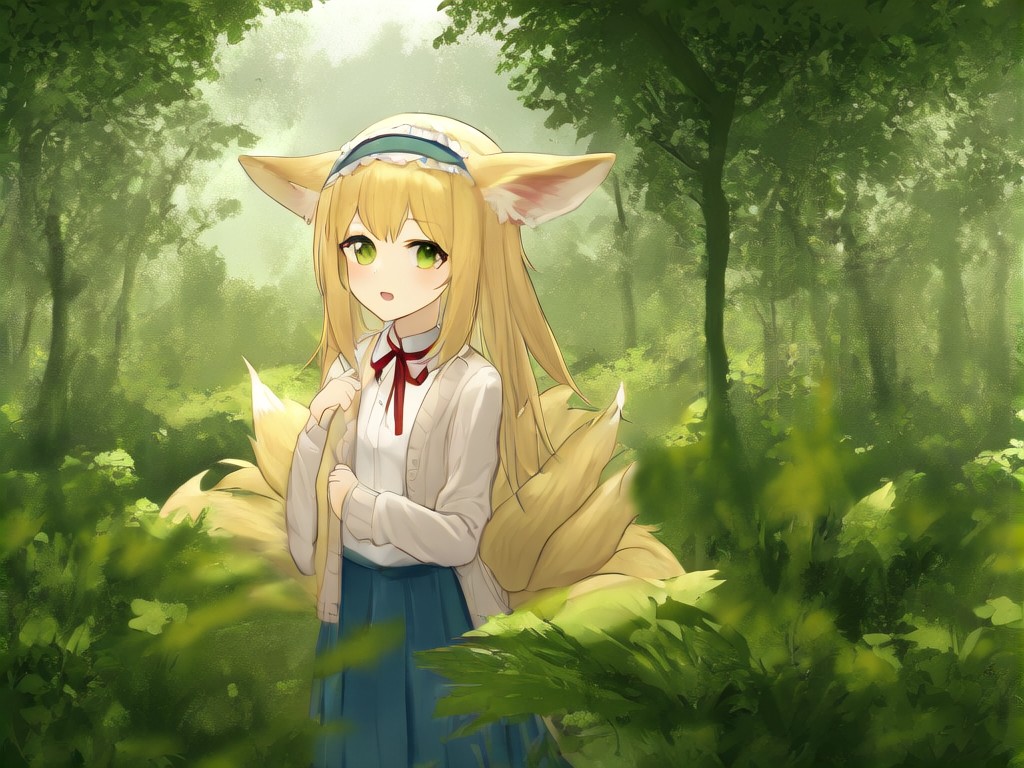} & 
\includegraphics[width=0.25\textwidth]{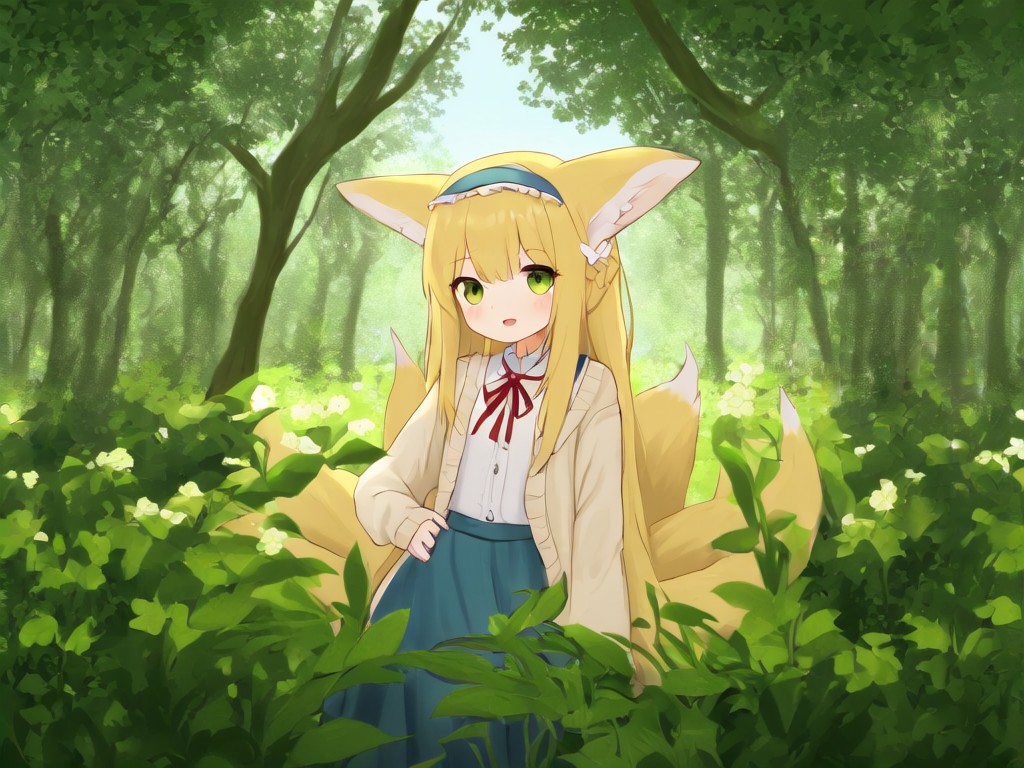} & 
\includegraphics[width=0.25\textwidth]{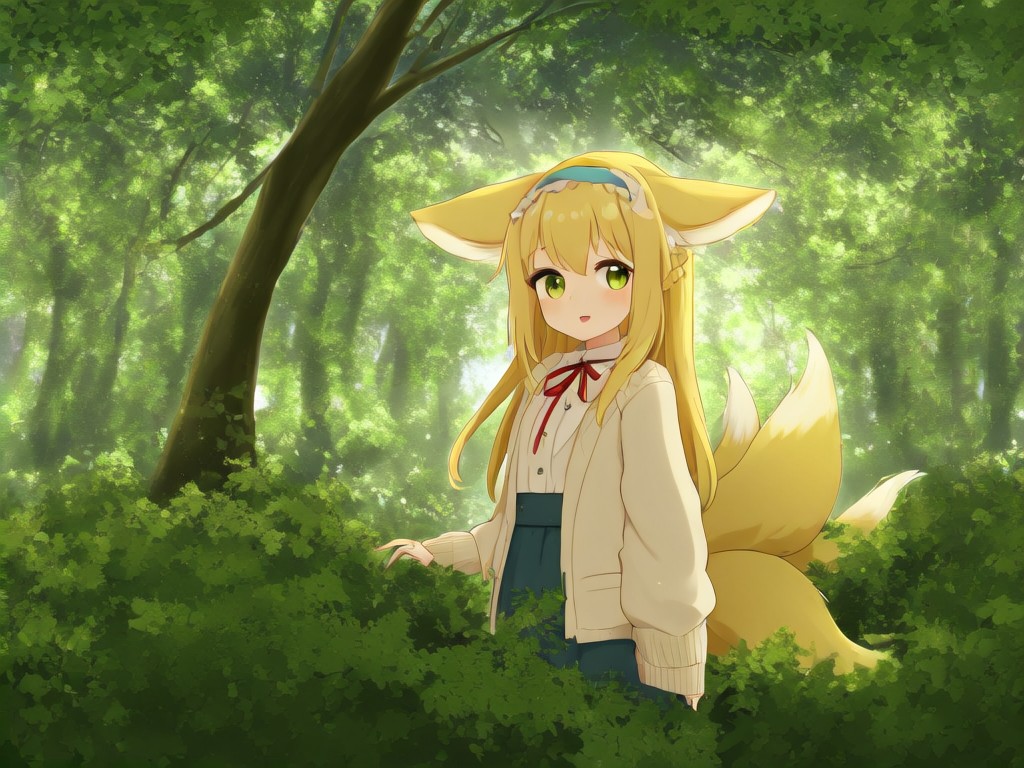} \\
\bottomrule
\end{tabular}
\caption{X-axis position map manipulation effects. Positive values shift the ``camera'' right, negative values shift left, enabling horizontal composition control.}
\label{tab:x_shift}
\end{table}

\begin{table}[h]
\centering
\begin{tabular}{lccc}
\toprule
\textbf{Y-Shift} & \textbf{+0.25 (Down)} & \textbf{No Shift} & \textbf{-0.25 (Up)} \\
\midrule
\textbf{Example} & \includegraphics[width=0.25\textwidth]{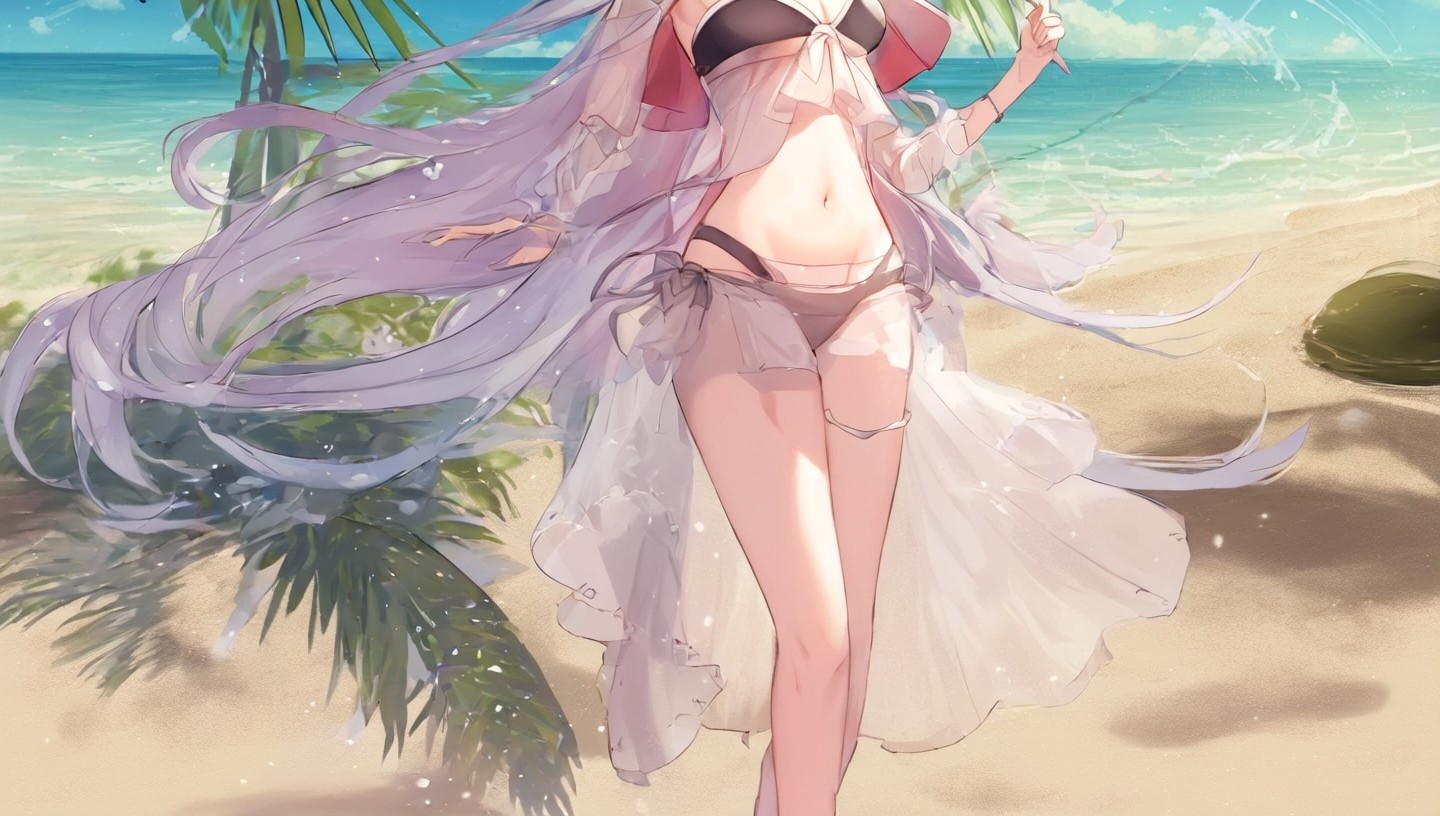} & 
\includegraphics[width=0.25\textwidth]{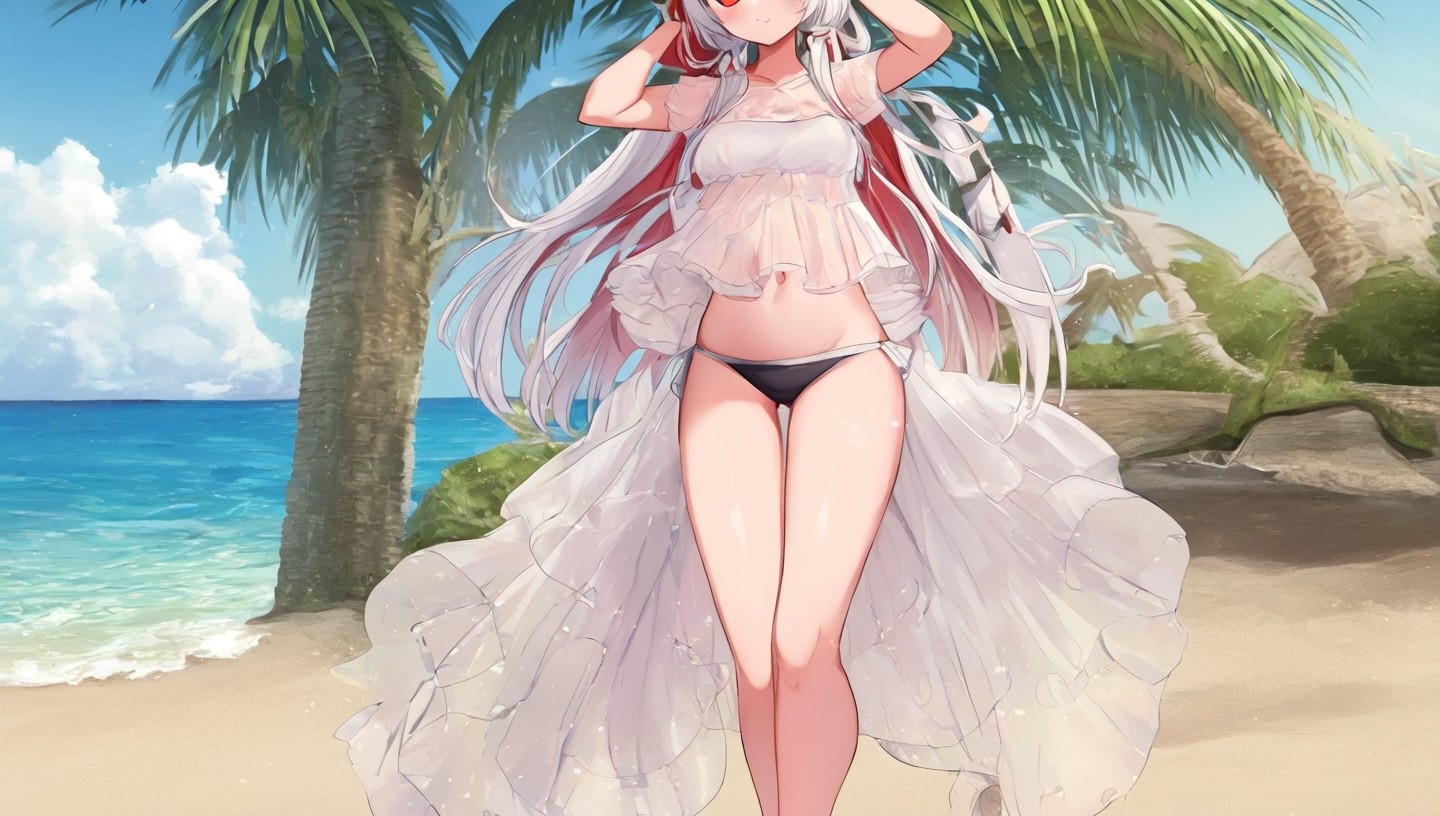} & 
\includegraphics[width=0.25\textwidth]{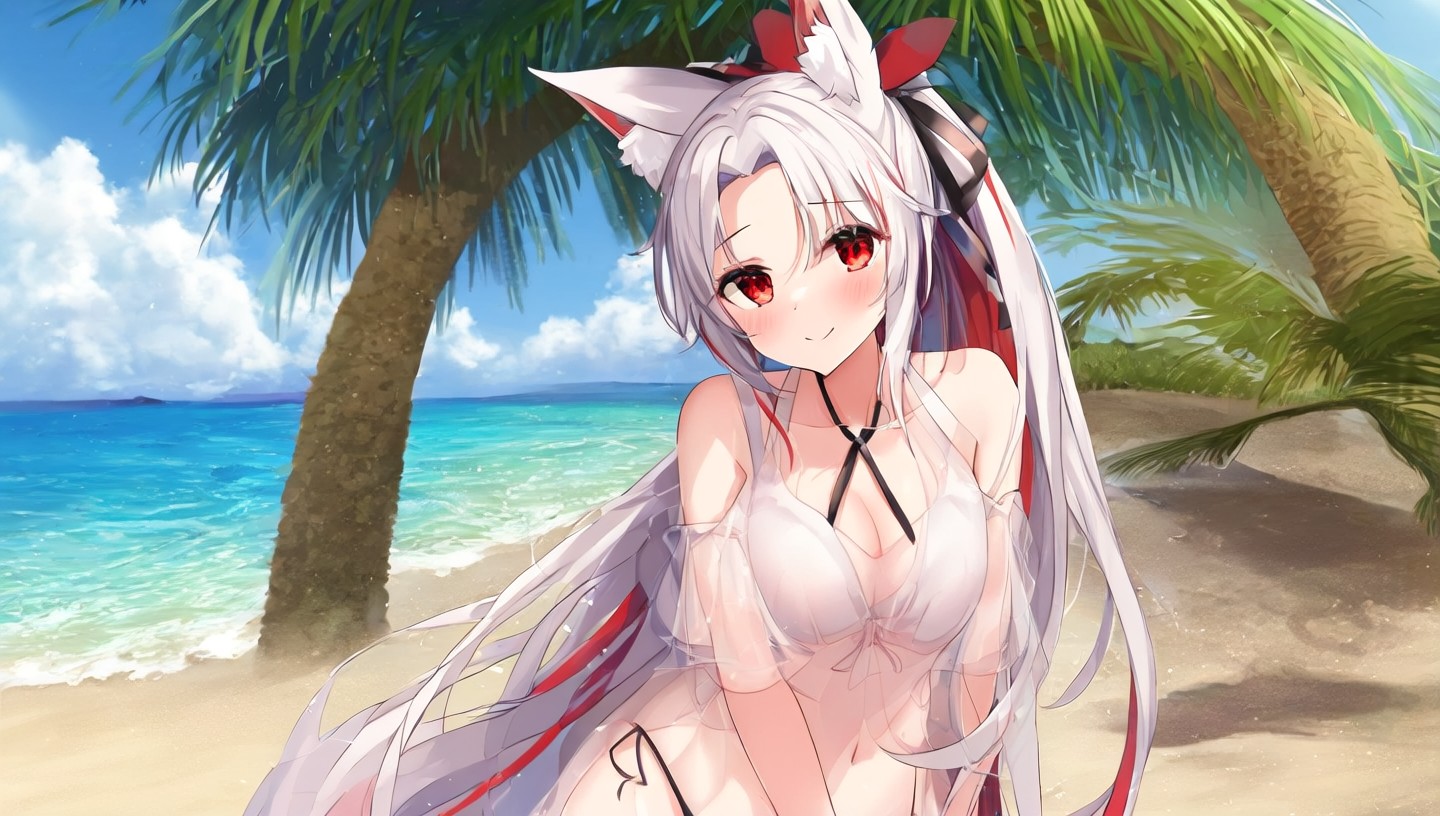} \\
\bottomrule
\end{tabular}
\caption{Y-axis position map manipulation effects. Positive values shift the ``camera'' down, negative values shift up, providing vertical composition control.}
\label{tab:y_shift}
\end{table}

\begin{table}[h]
\centering
\begin{tabular}{lccc}
\toprule
\textbf{Zoom} & \textbf{0.75 (Out)} & \textbf{No Zoom} & \textbf{1.33 (In)} \\
\midrule
\textbf{Example} & \includegraphics[width=0.25\textwidth,trim={0 400pt 0 200pt},clip]{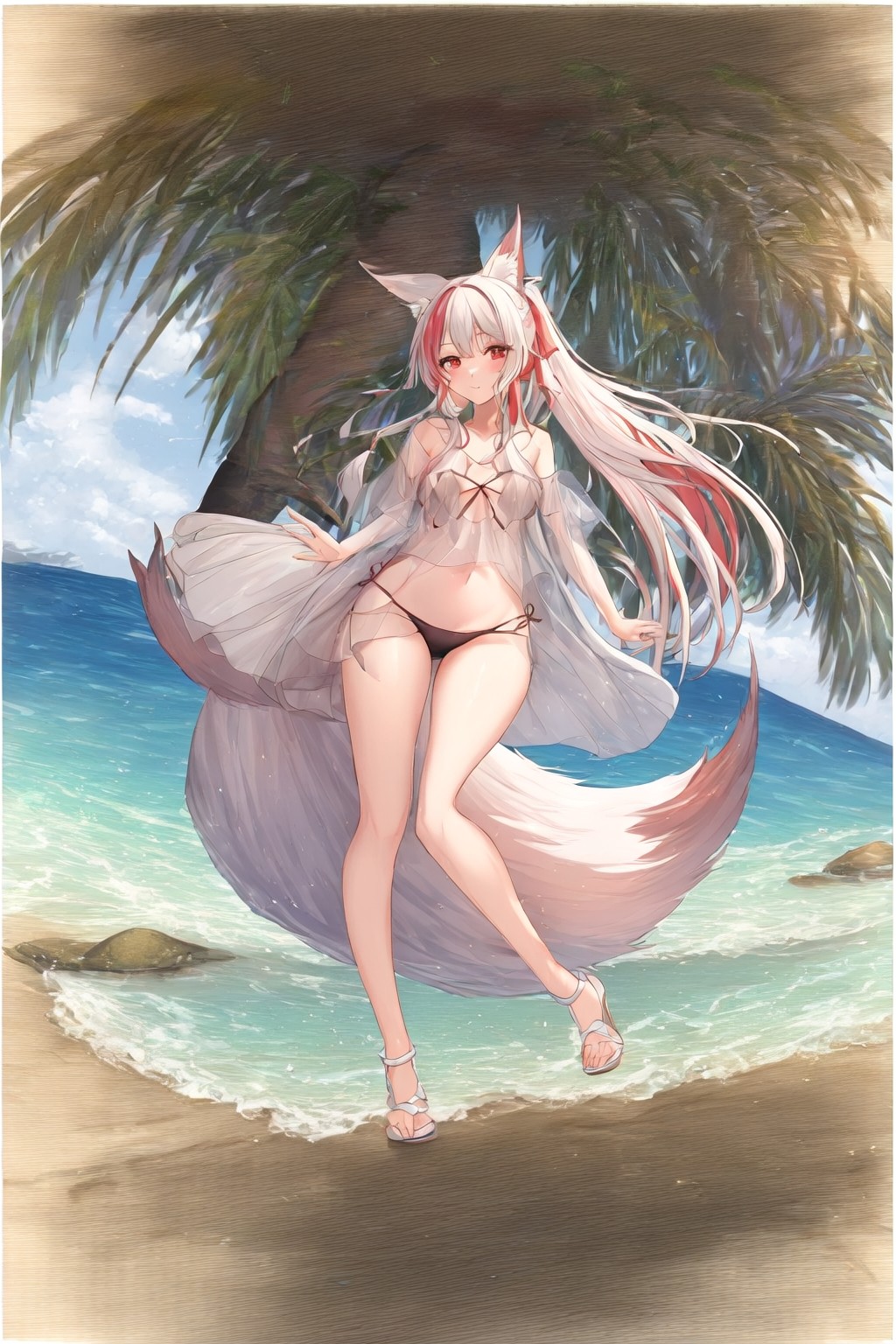} & 
\includegraphics[width=0.25\textwidth,trim={0 400pt 0 200pt},clip]{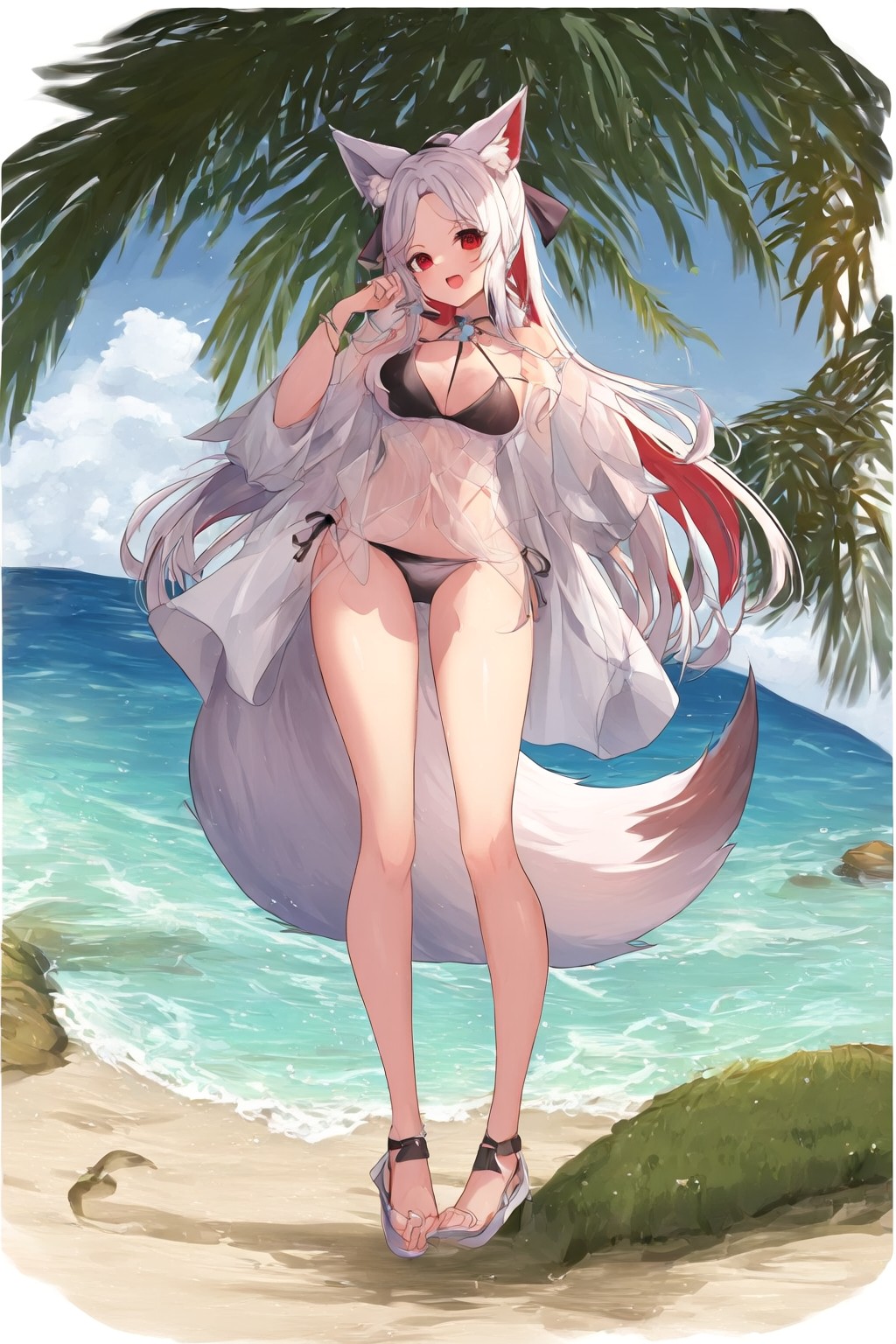} & 
\includegraphics[width=0.25\textwidth,trim={0 400pt 0 200pt},clip]{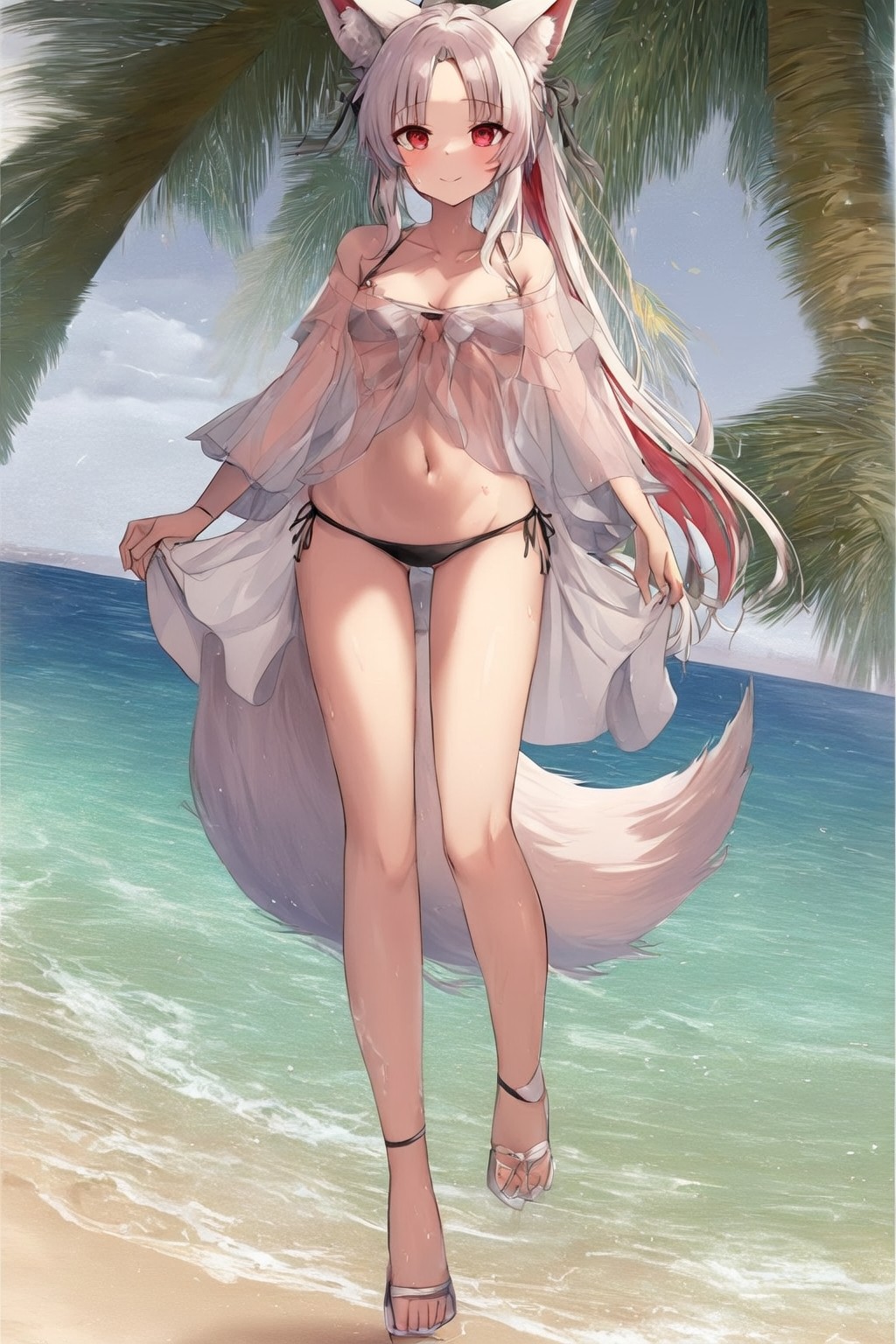} \\
\bottomrule
\end{tabular}
\caption{Zoom manipulation effects through position map scaling. Values less than 1.0 zoom out (showing more content), values greater than 1.0 zoom in (showing less content with higher detail).}
\label{tab:zoom}
\end{table}

These examples illustrate the camera-like control achievable through position map manipulation, providing users with intuitive spatial control over generation without requiring additional training or conditioning mechanisms. This capability emerges naturally from our shifted square crop training strategy and demonstrates the effectiveness of our positional encoding approach.

\subsection{Analysis and Discussion}

\textbf{Composition Consistency Mechanism:} The observed composition consistency likely stems from the combination of our XUT architecture's cross-attention skip connections and the training methodology. The cross-attention mechanism enables the model to maintain consistent spatial relationships while allowing fine-grained attribute modifications.

\textbf{Position Map Control Emergence:} The camera control capabilities emerge naturally from training with randomly cropped images combined with our positional encoding strategy. During training, the model learns to associate position maps with spatial layouts, enabling intuitive control during inference.

\textbf{Efficiency vs. Quality Trade-offs:} Despite the significantly smaller model size (343M parameters in XUT backbone vs. 2-8B in comparable models), HDM achieves competitive generation quality through architectural innovations and training optimizations. The composition consistency and spatial control capabilities suggest that careful architectural design can compensate for reduced parameter count.

\textbf{Consumer Hardware Viability:} The successful training on consumer hardware (4$\times$ RTX5090) demonstrates the viability of democratizing high-quality text-to-image model development. The total training cost of \$535-620 represents a dramatic reduction from traditional approaches, making advanced diffusion research accessible to individual researchers and smaller organizations.

%% file: sections/conclusion.tex
\section{Conclusion}
\label{sec:conclusion}

We have presented HDM (Home-made Diffusion Model), a novel approach to efficient text-to-image generation that demonstrates competitive performance while dramatically reducing computational requirements for training. Our key contributions include the XUT (Cross-U-Transformer) architecture, which employs cross-attention for skip connections in U-shaped transformers, and a comprehensive training recipe that enables high-quality 1024$\times$1024 generation using consumer-level hardware.

The successful training of HDM on four RTX5090 GPUs at a cost of \$535-620 represents a paradigm shift in accessibility for diffusion model research. By achieving competitive results with a 343M parameter backbone (compared to 2-8B parameters in comparable models), we demonstrate that architectural efficiency and training optimization can effectively compensate for reduced model scale.

Our experimental results reveal several notable capabilities: (1) composition consistency under prompt modifications, enabling fine-grained attribute control while preserving overall image structure; (2) emergent camera control through position map manipulation, providing intuitive spatial control without additional training; and (3) efficient inference with flexible aspect ratio support while the training pipeline only need to handle square crops.

The HDM framework challenges the prevailing assumption that high-quality text-to-image generation requires massive computational resources, potentially democratizing access to advanced generative modeling research. By demonstrating that state-of-the-art results can be achieved on consumer hardware, we open new possibilities for individual researchers and smaller organizations to contribute to the field.

\textbf{Future Directions:} Our ongoing work focuses on scaling to larger, more diverse datasets (targeting 40M+ images), exploring both larger and smaller model configurations, and investigating pixel-space approaches. We also plan to develop ``Hires Fix'' or refiner approaches for enhanced quality at higher resolutions.

The HDM project represents a step toward more accessible and democratized AI research, where innovation is not limited by computational resources but driven by architectural creativity and training efficiency. We believe this approach will accelerate progress in text-to-image generation by lowering barriers to participation and enabling a broader community of researchers to contribute to the field.

%% file: sections/appendix.tex
\clearpage
\appendix
\clearpage

\noindent\rule{\textwidth}{1pt}
\begin{center}
\vspace{3pt}
{\Large Appendix}
\vspace{-3pt}
\end{center}
\noindent\rule{\textwidth}{1pt}

\section*{Table of Contents}

\definecolor{sectionblue}{RGB}{65, 105, 225}
\titlecontents{section}
  [1.6em]                                           
  {\addvspace{0.5pc}\color{sectionblue}}          
  {\contentslabel[\thecontentslabel]{1.75em}}      
  {\hspace*{-1.5em}}                              
  {\hfill\contentspage}                           
  []                                              

\titlecontents{subsection}
  [4em]                                         
  {\addvspace{0.2pc}\color{sectionblue}}          
  {\contentslabel[\thecontentslabel]{2.5em}}      
  {\hspace*{-2.5em}}                              
  {\titlerule*[1pc]{.}\contentspage}              
  []                                              

\startcontents[appendix]
\printcontents[appendix]{l}{1}{\setcounter{tocdepth}{2}}

\newpage

\section{Implementation Details}
\label{app:implementation}

In this section, we provide PyTorch/Python-style pseudocode for key components in training HDM models, detailing the mechanisms introduced in Section~\ref{sec:methodology}

\subsection{Position Map Generation}
\label{app:pos_map_impl}

Following pseudo code provide a reference implementation of the position map strategy we used in HDM.

\begin{minted}{python}
def generate_position_map(latent_tensor):
    _c, h, w = latent_tensor.shape
    aspect_ratio = h / w
    h_range = aspect_ratio**0.5
    w_range = 1 / h_range

    h_coords = linspace(-h_range, h_range, h)
    w_coords = linspace(-w_range, w_range, w)

    position_map = stack(meshgrid(h_coords, w_coords, indexing="ij"))
    return position_map
\end{minted}
\vspace{2em}

\subsection{Shifted Square Crop Strategy}
\label{app:shifted_square_crop_impl}
Following pseudo code provide a reference implementation of the "shorter edge resizing" strategy.

\begin{minted}{python}
def resize_image_min_dim(image_object, target_min_dim):
    W, H = image_object.size
    scale_factor = target_min_dim / min(H, W)
    
    if H > W:
        scaled_H = int(H * scale_factor)
        return image_object.resize((target_min_dim, scaled_H))
    else:
        scaled_W = int(W * scale_factor)
        return image_object.resize((scaled_W, target_min_dim))
\end{minted}
\vspace{2em}

Following pseudo code provide a reference implementation of the "shifted square crop" strategy.

\begin{minted}{python}
def apply_shifted_square_crop(image, pos_map):
    _c, h_resized, w_resized = resized_image_tensor.shape

    if h_resized > w_resized:
        index_h = randint(0, h_resized - w_resized)
        image_cropped = image[:, index_h : index_h + w_resized, :]
        pos_map_cropped = pos_map[:, index_h : index_h + w_resized, :]
    else: # w_resized >= h_resized
        index_w = randint(0, w_resized - h_resized)
        image_cropped = image[:, :, index_w : index_w + h_resized]
        pos_map_cropped = pos_map[:, :, index_w : index_w + h_resized]

    return image_cropped, pos_map_cropped
\end{minted}
\vspace{2em}

\newpage
\section{Image Examples}
\label{app:examples}

In this sections we provide more image examples generated from HDM-XUT-base model.

\import{sections/}{gallery.tex}

\newpage

\section{Limitations}
While this report demonstrates the viability of training a high-quality text-to-image model on consumer hardware, we acknowledge several limitations. The primary constraint of this work is the lack of comprehensive quantitative evaluation. Due to resource limitations, we have not conducted extensive ablation studies or benchmarked our model against established metrics. Consequently, while the performance of the proposed XUT architecture is empirically strong, we cannot definitively claim its superiority over alternative U-shaped transformer designs without a rigorous comparative analysis.
Furthermore, the HDM framework integrates several individually validated techniques, such as TREAD acceleration and EQ-VAE fine-tuning. However, the synergistic effects and interdependencies of this specific combination remain unexplored. The primary contribution of this report should therefore be viewed as an existence proof: it demonstrates that a competitive text-to-image model can be successfully trained within a highly constrained budget. We do not claim that the presented configuration is optimal, but rather that it represents a viable and effective pathway for democratized research in this domain.

\section{Future Work}
Building upon the foundation established in this report, our future work will proceed along several key directions. First, to address the current limitations, we plan to establish a comprehensive benchmarking pipeline specifically designed for low-budget text-to-image training. This will allow us to systematically validate the architectural and training choices made in the HDM project and perform the necessary ablation studies to isolate the impact of each component.
Second, we aim to assess the generalization capabilities of the HDM framework by training it on broader, more general-purpose datasets. While the Danbooru2023 dataset provided a rich domain for initial validation, training on large-scale public datasets such as CC12M or COYO-700M will be crucial for proving the model's broader applicability. Concurrently, we will train the different model scales (XUT-small and XUT-large) outlined in Table~\ref{tab:model_specs} to analyze the performance and efficiency trade-offs across various parameter counts.
Finally, we will continue to explore component-level optimizations and investigate the framework's robustness under more extreme conditions. This includes exploring alternative autoencoder architectures that may offer faster convergence or more favorable latent space characteristics. We also intend to study the model's behavior in highly constrained scenarios, such as training on limited dataset sizes, to further understand the boundaries of efficient text-to-image generation.

\section{Acknowledgements}
\label{app:acknowledgements}

We extend our sincere gratitude to the research community and individuals who contributed to the development of HDM:

\textbf{Technical Contributions}: Felix Krause and Stefan Andreas Baumann (TREAD authors) provided invaluable insights and assistance during HDM development, particularly regarding efficient token routing integration and inference optimization.

\textbf{Community Insights}: AngelBottomless (Illustrious series model author), Mahouko (Birch Lab), and Uptightmoose provided valuable insights throughout the HDM planning process, contributing to architectural decisions and training strategy development.

\textbf{Open Source Community}: The broader open-source machine learning community, whose tools, libraries, and shared knowledge made this consumer-level training approach possible.

%% file: sections/gallery.tex

\begin{table}[htbp]
\centering
\begin{tabular}{|>{\centering\arraybackslash}p{0.08\textwidth}|>{\centering\arraybackslash}p{0.86\textwidth}|}
\hline
\textbf{Image} & \includegraphics[width=0.85\textwidth,height=0.6\textheight,keepaspectratio]{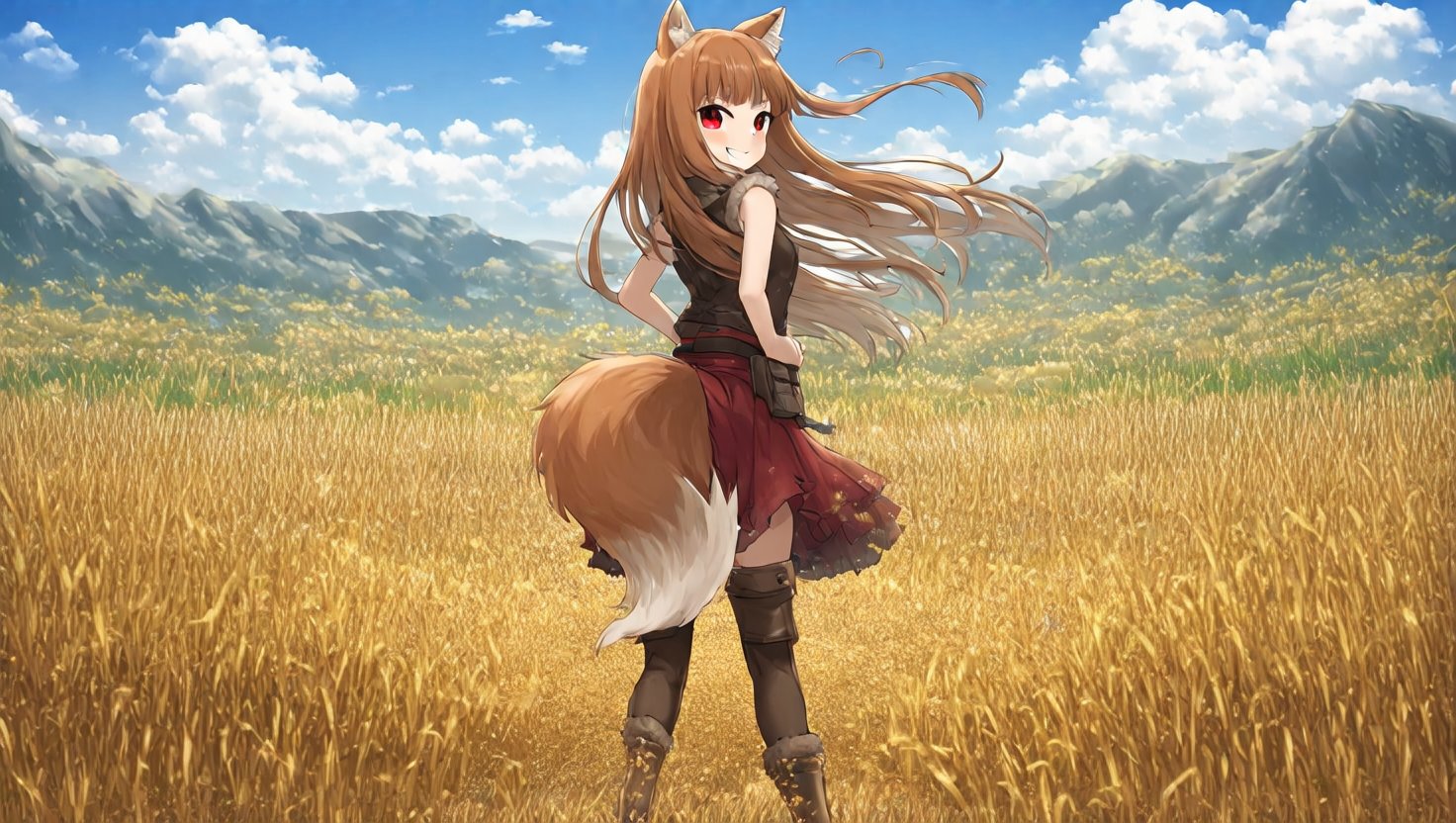} \\
\hline
\textbf{Prompt} & \begin{lstlisting}
1girl, holo, spice and wolf, shoichi \(ekakijin\), full body, anniversary, sky, outdoors, wheat, wolf ears, brown hair, smile, fur trim, sleeveless, sun glare, long hair, solo, boots, sunlight, tail, animal ears, pouch, mountainous horizon, cart, wolf tail, wolf girl, sleeveless jacket, skirt, day, scenery, nature, blush, blue sky, mountain, standing, brown footwear, looking at viewer, jacket, wind lift, red eyes, wind, wheat field, very long hair, fur-trimmed boots, cloud, cloudy sky, hand on own hip, thigh boots, looking back, forest, floating island, red skirt, fur-trimmed jacket, floating city, from behind, floating hair, thighhighs,

In this vibrant and whimsical artwork by the artist shoichi (ekakijin), a character from the series "Spice and Wolf" stands prominently in a field of golden wheat. The central figure, Holo, is depicted with her signature long brown hair cascading down her back, and she sports wolf-like ears and a tail, adding an element of fantasy to her appearance. Her red eyes sparkle as she looks back over her shoulder, smiling warmly at the viewer. The sky above is a clear blue, dotted with fluffy white clouds, creating a serene backdrop for this enchanting scene.

masterpiece, newest, official art, commentary request, absurdres 
\end{lstlisting} \\
\hline
\end{tabular}
\end{table}

\begin{table}[htbp]
\centering
\begin{tabular}{|>{\centering\arraybackslash}p{0.08\textwidth}|>{\centering\arraybackslash}p{0.86\textwidth}|}
\hline
\textbf{Image} & \includegraphics[width=0.85\textwidth,height=0.6\textheight,keepaspectratio]{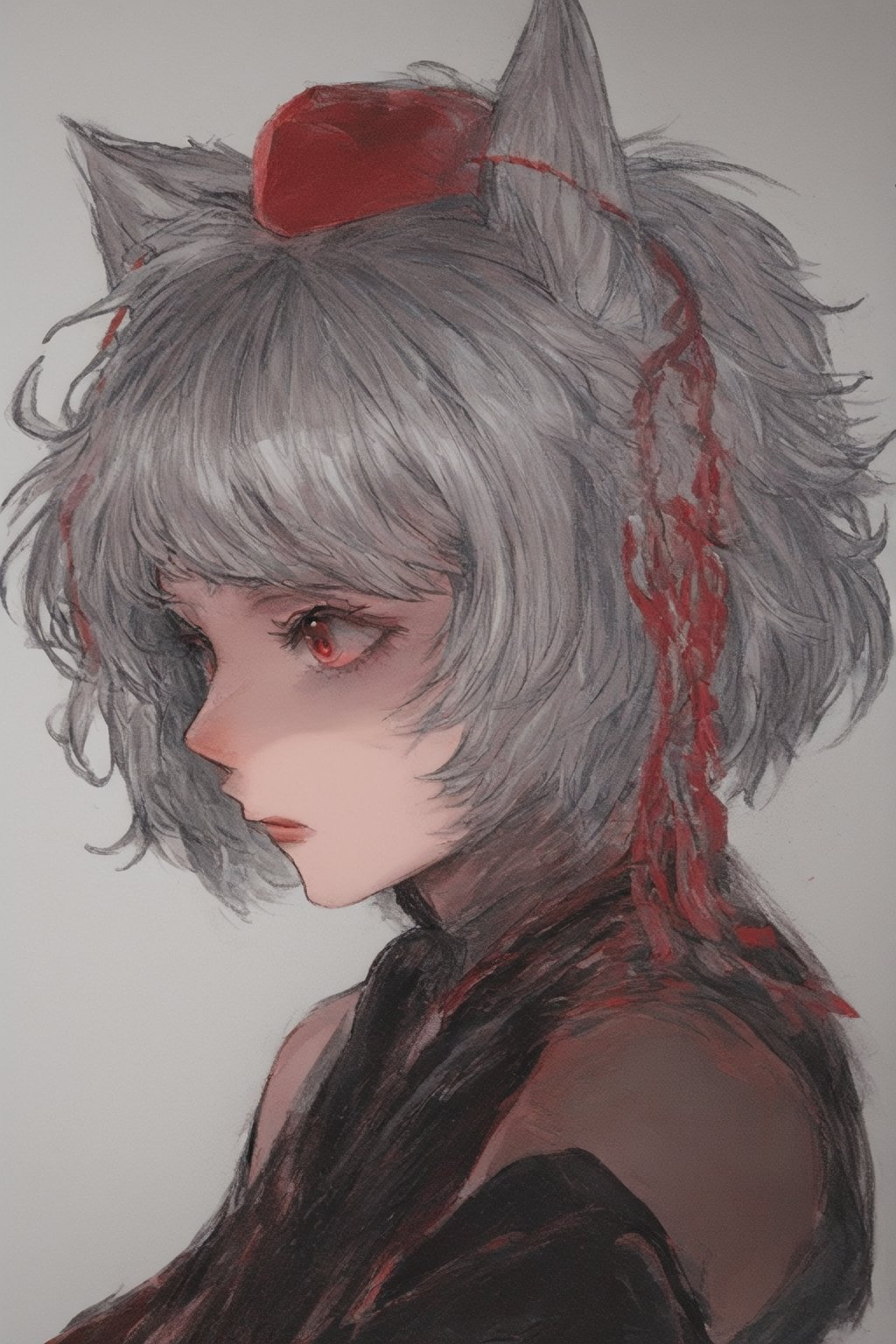} \\
\hline
\textbf{Prompt} & \begin{lstlisting}
1girl, inubashiri momiji, touhou, matsuda \(matsukichi\), 

red eyes, profile, grey hair, upper body, tassel, painterly, japanese clothes, animal ears, short hair, wolf ears, tokin hat, detached sleeves, hat, simple background, solo, pom pom \(clothes\), close-up, messy hair, closed mouth, expressionless, breasts, character name, eyelashes, from side, faux traditional media, white background, nose, turtleneck, headphone, medium breasts, looking away, shaded face, bob cut, lips, bare shoulders,

This image is a striking piece of digital art by the artist Matsuda (Matsukichi), known for their distinctive style within the Touhou series. The central figure in this portrait is Inubashiri Momiji, depicted with her signature features including short grey hair adorned with a red tassel and striking red eyes. She wears traditional Japanese attire, including a tokin hat that adds an air of mystery to her appearance. Her expression is serious and introspective as she gazes off to the side, creating a sense of depth and intrigue in the composition.

masterpiece, newest, photoshop \(medium\), commentary request, absurdres 
\end{lstlisting} \\
\hline
\end{tabular}
\end{table}

\begin{table}[htbp]
\centering
\begin{tabular}{|>{\centering\arraybackslash}p{0.08\textwidth}|>{\centering\arraybackslash}p{0.86\textwidth}|}
\hline
\textbf{Image} & \includegraphics[width=0.85\textwidth,height=0.6\textheight,keepaspectratio]{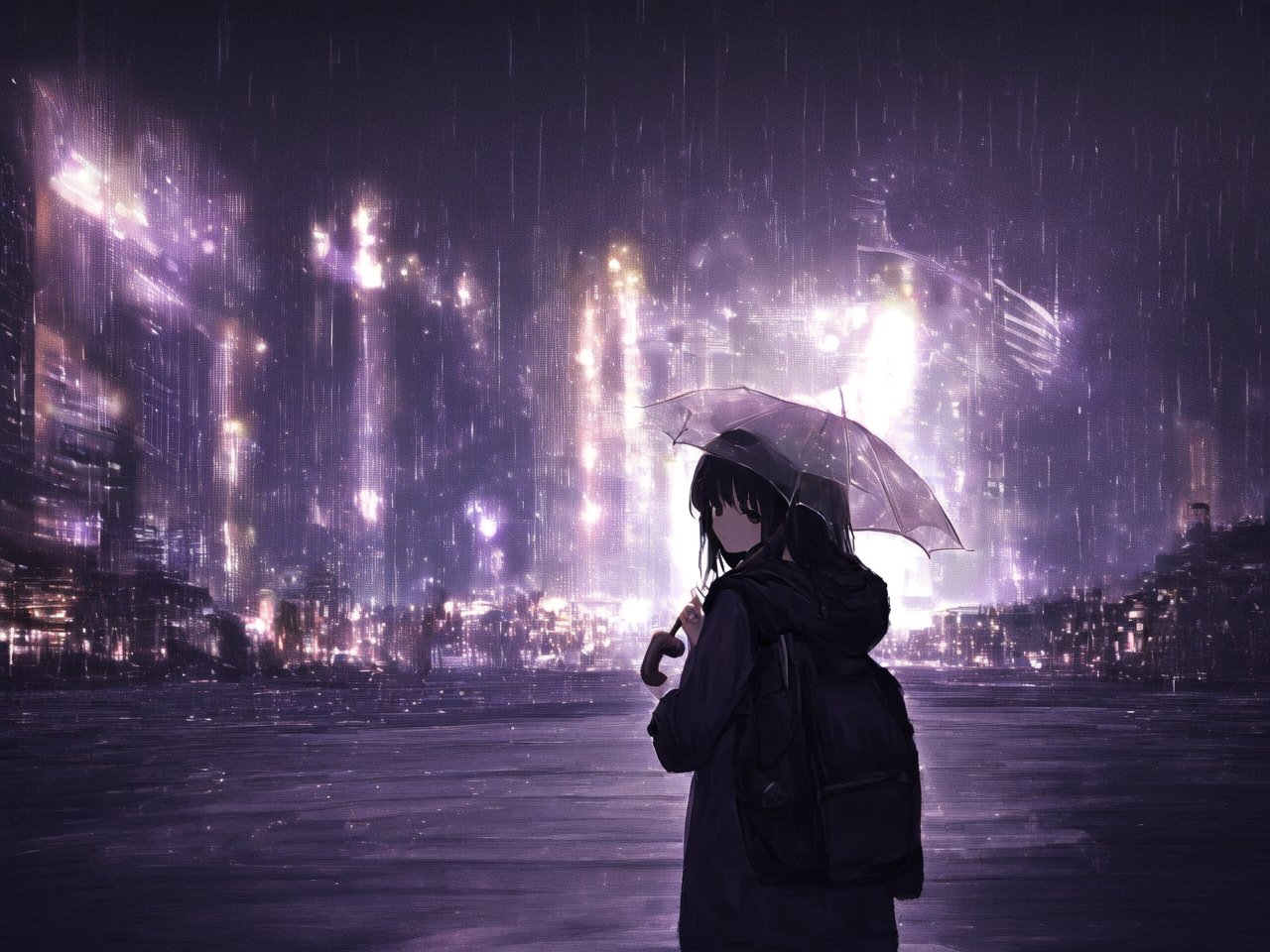} \\
\hline
\textbf{Prompt} & \begin{lstlisting}
1girl, original, catzz, 

people, road, neon palette, city, wet, blurry, coat, limited palette, long hair, bag, umbrella, night, rain, lights, black hair, transparent umbrella, transparent, long sleeves, arm at side, looking back, scenery, blurry foreground, depth of field, dark, backpack, outdoors, crosswalk, looking at viewer, english text, street, hooded coat, bokeh, city lights, solo focus, from behind, standing, building, hood down, hood, purple theme, cowboy shot, black eyes, holding, puddle, holding umbrella,

The image by the artist catzz depicts a solitary figure standing on a rain-soaked street at night, illuminated by neon lights. She holds an umbrella in her right hand, which is raised as if caught mid-stride. Her posture suggests she is walking along the crosswalk, with the wet pavement reflecting the glow of distant traffic signals and the city's ambient light. The overall atmosphere is one of solitude and quiet introspection amidst the urban landscape.

masterpiece, newest, mixed-language commentary, commentary, absurdres 
\end{lstlisting} \\
\hline
\end{tabular}
\end{table}

\begin{table}[htbp]
\centering
\begin{tabular}{|>{\centering\arraybackslash}p{0.08\textwidth}|>{\centering\arraybackslash}p{0.86\textwidth}|}
\hline
\textbf{Image} & \includegraphics[width=0.85\textwidth,height=0.6\textheight,keepaspectratio]{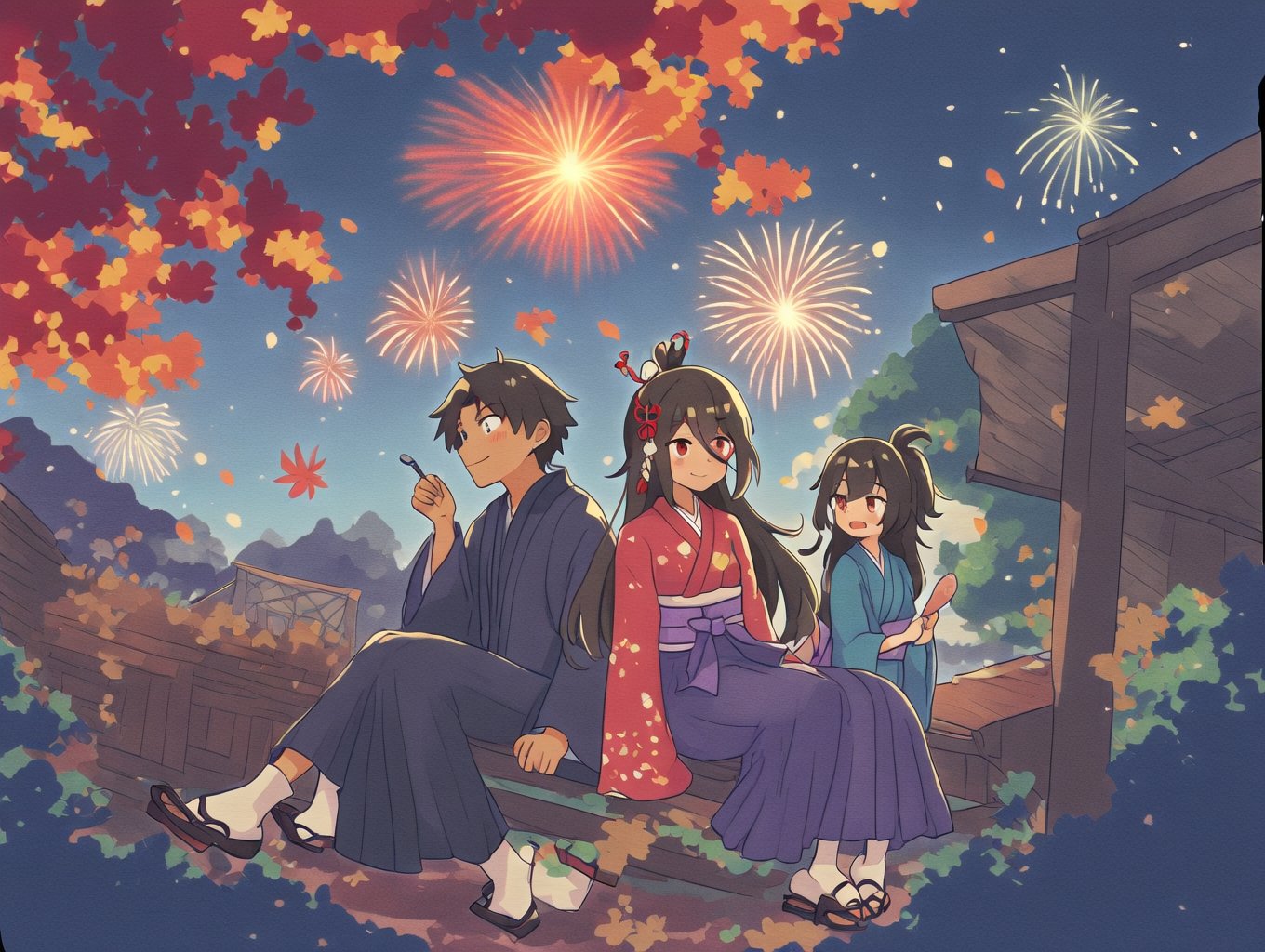} \\
\hline
\textbf{Prompt} & \begin{lstlisting}
1girl, 1girl, yamada tae, ajna \(indivisible\), zombie land saga, indivisible, igarashi tsukiyo \(tukiyo320\),  scenery, red eyes, hair between eyes, black hair, very long hair, brown eyes, short hair, brown hair, hair ornament, igarashi tsukiyo, (tukiyo320:1.1), painting \(action\), medium hair, night, sitting, kanzashi, geta, haori himo, sash, blush, holding paintbrush, obi, looking at another, holding, light smile, long skirt, fireworks print, sky, tentacle hair, paintbrush, treehouse, closed mouth, skirt, long hair, faux traditional media, japanese clothes, cloud, alternate form, sitting in window, purple sash, squidbeak splatoon, dark-skinned female, dark skin, red kimono, blue kimono, best qualiytg, In the image by igarashi tsukiyo (sdrsyou320), known for their distinctive and whimsical style, we see a captivating scene from the series "Zombie Land Saga" inspired by Studio Ghibli. The composition centers on three main characters perched on treehouses within a magical forest backdrop adorned with autumn leaves in hues of orange and red. The artist has utilized the fine quality to depict these whimsically dressed figures amidst the natural, mystical setting, capturing both fantasy and cultural elements in this enchanting scene from the "Indivisest" series within the franchise. masterpiece, great quality, sfw Negative prompt: unknownlow quality, worst quality, text, signature, jpeg artifacts, bad anatomy, old, early, copyright name, watermark, artist name, signature, weibo username, mosaic censoring, bar censor, censored, text, speech bubbles, doll, character doll, hair intake, realistic, 2girls, 3girls, multiple girls, crop top, cropped head, cropped, large breasts, nsfw Steps: 35, Sampler: euler_simple, CFG Scale: 1.0, Seed: 960573175128065, Size: 1360x1024, Model hash: f7e7841239, Model: hdm-xut-340M-1024px-9kstep-note, Version: ComfyU 
\end{lstlisting} \\
\hline
\end{tabular}
\end{table}

\begin{table}[htbp]
\centering
\begin{tabular}{|>{\centering\arraybackslash}p{0.08\textwidth}|>{\centering\arraybackslash}p{0.86\textwidth}|}
\hline
\textbf{Image} & \includegraphics[width=0.85\textwidth,height=0.6\textheight,keepaspectratio]{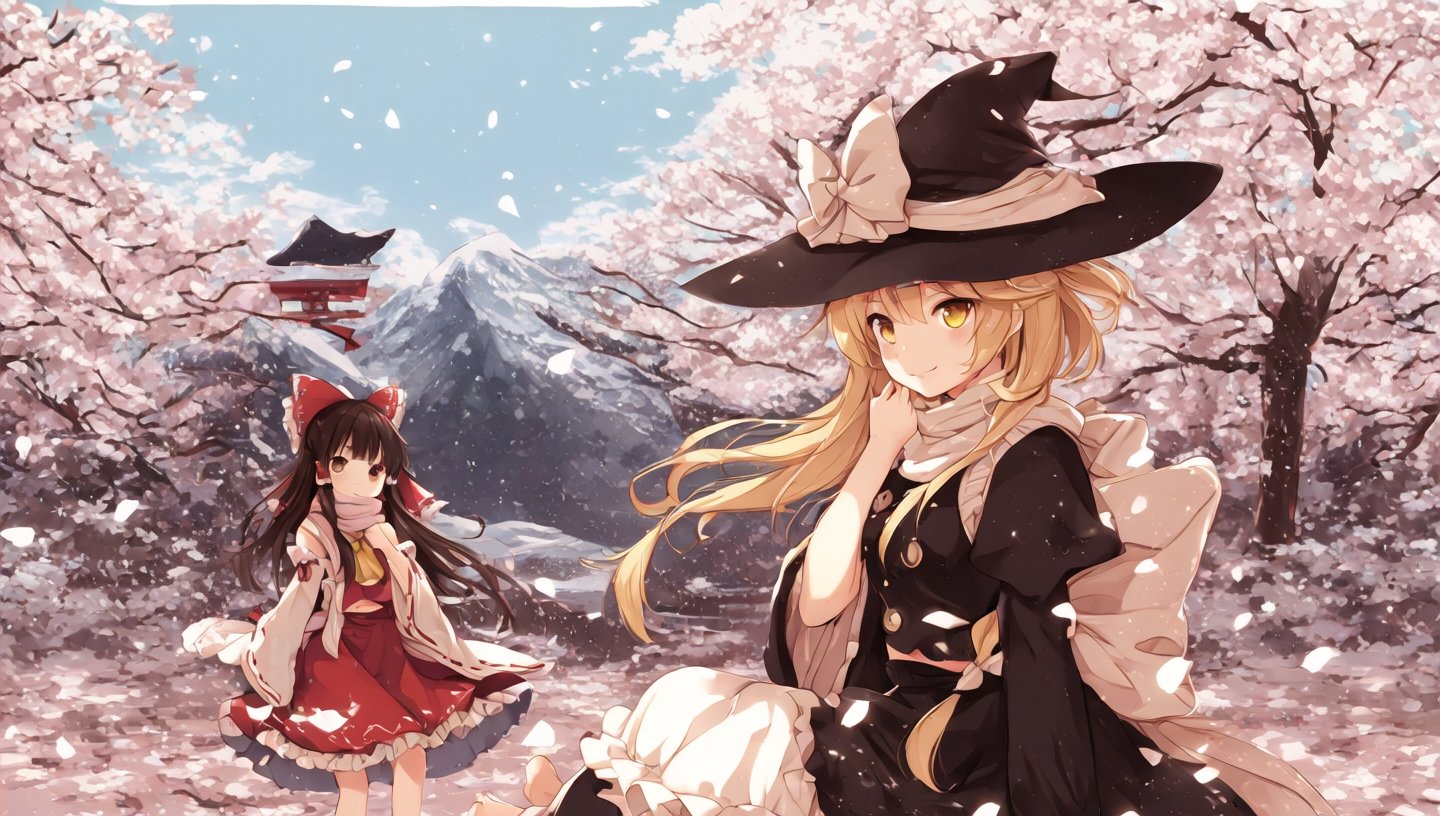} \\
\hline
\textbf{Prompt} & \begin{lstlisting}
multiple girls, 2girls, kirisame marisa, hakurei reimu, hidden star in four seasons, touhou, ke-ta, 

bow, black hat, wide sleeves, tree, scarf, juliet sleeves, scenery, blunt bangs, shared clothes, torii, skirt, barefoot, yellow ascot, white apron, midriff peek, ascot, mountain, ribbon trim, snow, skirt set, puffy sleeves, red skirt, pink scarf, shared scarf, petals, hair bow, black dress, sky, waist apron, sitting, detached sleeves, outdoors, hat, hair tubes, hand up, white bow, red bow, long sleeves, wide shot, witch hat, ribbon-trimmed sleeves, dress, petticoat, blonde hair, looking at viewer, day, brown hair, sweater, brown eyes, sweater dress, apron, cherry blossoms, sitting on torii, snowflake print, floral print, hat bow, long hair,

This image is a beautifully detailed illustration by the artist ke-ta, known for their intricate and expressive style. The artwork features two characters from the series "Hidden Star in Four Seasons" of the Touhou project, set against a picturesque backdrop of cherry blossom trees and snowy mountains. The scene is serene yet captivating, with the cherry blossoms adding a touch of natural beauty to the composition. The characters' interactions and the harmonious setting create a sense of camaraderie and tranquility, showcasing ke-ta's skill in character design and storytelling within the Touhou universe.

masterpiece, newest, doujinshi, scan, translation request, absurdres 
\end{lstlisting} \\
\hline
\end{tabular}
\end{table}

\begin{table}[htbp]
\centering
\begin{tabular}{|>{\centering\arraybackslash}p{0.08\textwidth}|>{\centering\arraybackslash}p{0.86\textwidth}|}
\hline
\textbf{Image} & \includegraphics[width=0.85\textwidth,height=0.6\textheight,keepaspectratio]{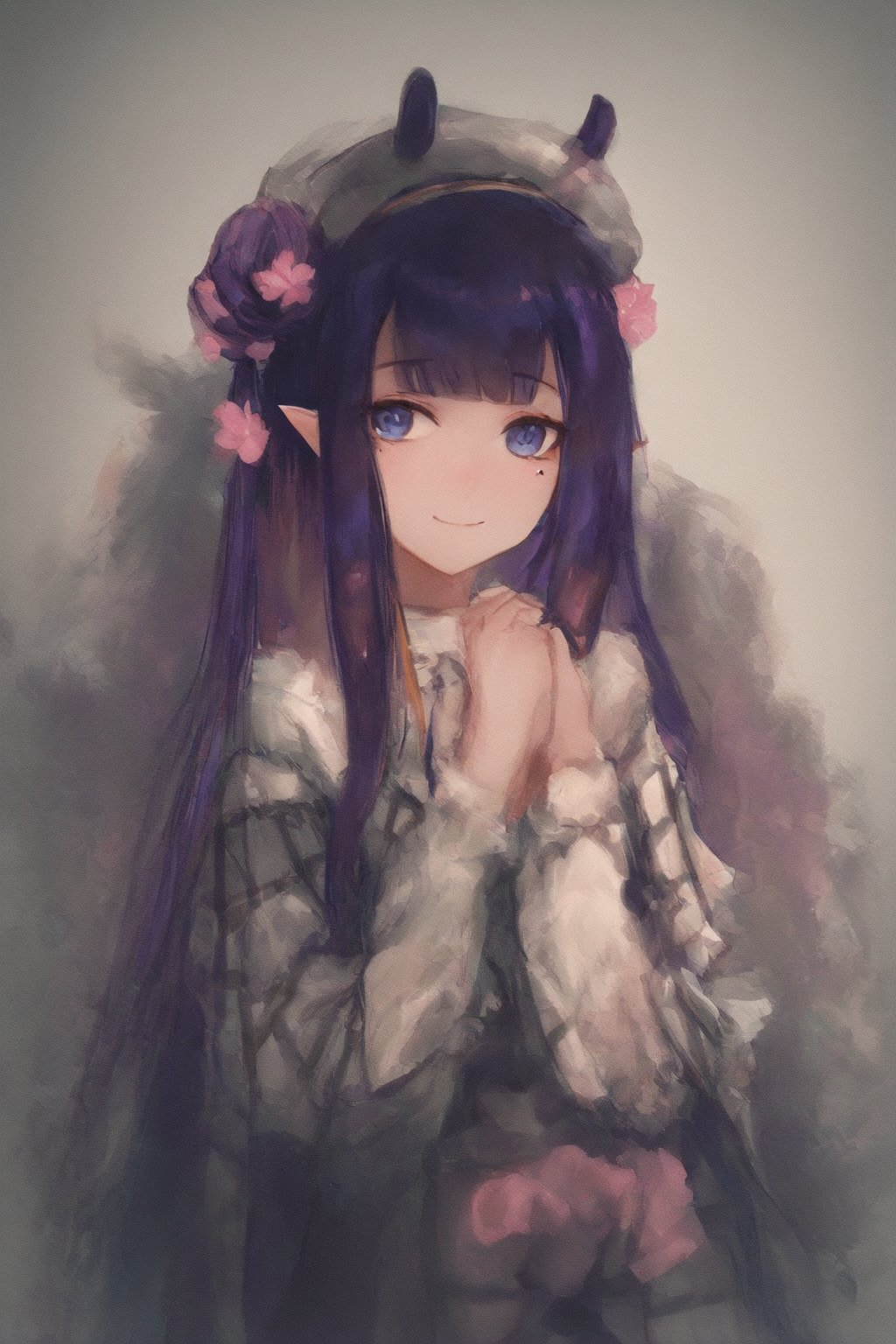} \\
\hline
\textbf{Prompt} & \begin{lstlisting}
1girl, ninomae ina'nis, hololive english, hololive, ninomae ina'nis \(artist\), 

branch, own hands together, blunt bangs, pointy ears, very long hair, long hair, gradient hair, dress, tentacle hair, beret, purple hair, dated, white dress, two-tone hair, animal ears, upper body, signature, pink flower, hair bun, mole, own hands clasped, single side bun, light smile, hat, sidelocks, multicolored hair, virtual youtuber, mole under eye, straight hair, orange hair, smile, plaid, flower, solo, alternate costume, blue eyes, looking at viewer, white hat, closed mouth, plaid dre,

a digital illustration by the artist ninomae ina'nis (artist), known for their distinctive and detailed style. The artwork features a character from the Hololive English series, specifically Ninomae Ina'nis, who is depicted with long, dark hair adorned with pink highlights and pointed ears. She wears an alternate costume that includes animal-like ears and a beret, adding to her whimsical appearance. The background is a soft gradient of light colors, subtly accentuating the character's features and creating a serene atmosphere.

masterpiece, newest, commentary, english commentary, self-portrait, absurdres 
\end{lstlisting} \\
\hline
\end{tabular}
\end{table}

\begin{table}[htbp]
\centering
\begin{tabular}{|>{\centering\arraybackslash}p{0.08\textwidth}|>{\centering\arraybackslash}p{0.86\textwidth}|}
\hline
\textbf{Image} & \includegraphics[width=0.85\textwidth,height=0.6\textheight,keepaspectratio]{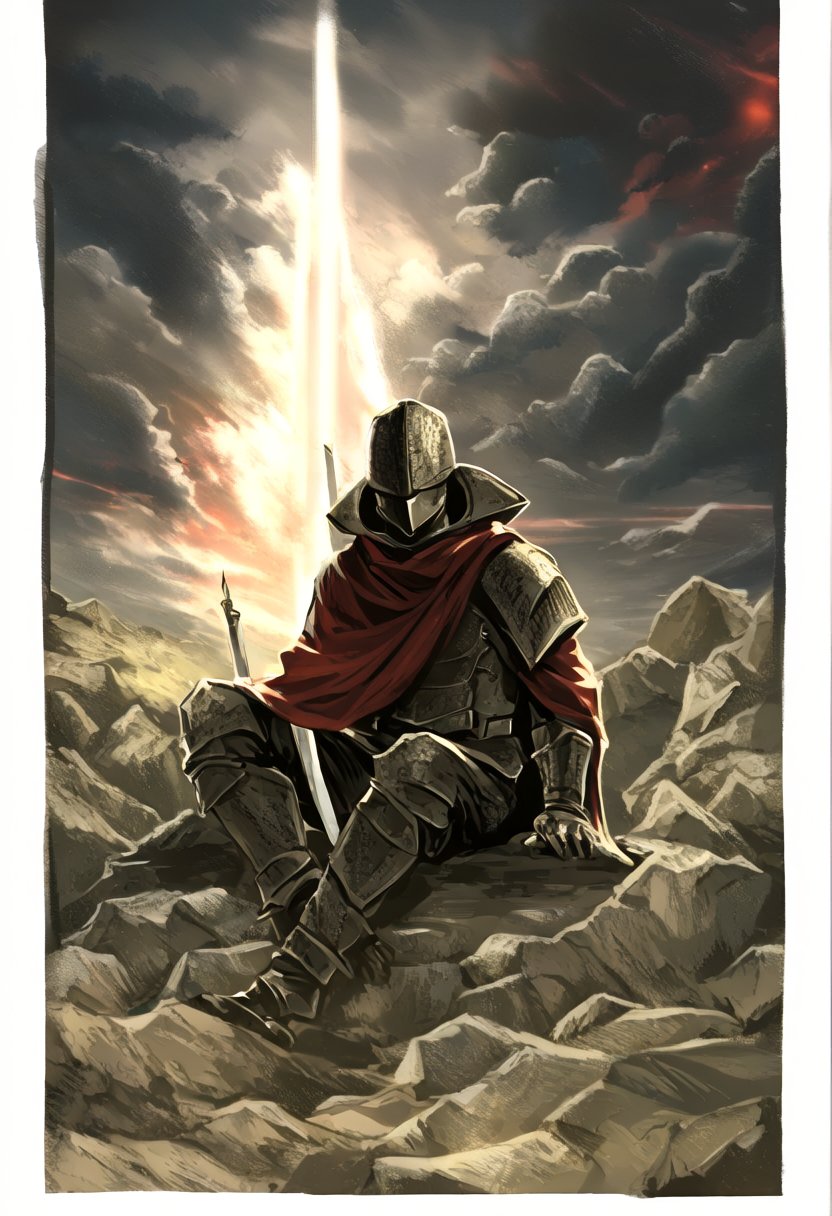} \\
\hline
\textbf{Prompt} & \begin{lstlisting}
1boy, 

year 2025, solo, knight, armor, full armor, plate armor, sitting, sword, weapon, cape, red cape, helmet, medieval, fantasy, rocks, outdoors, sky, clouds, dramatic, melancholy, somber, weathered, battle-worn, resting, contemplative, red light, beam of light, artistic, painting, textured, gritty, atmospheric, moody, warrior, crusader, templar, chainmail, metal armor, exhausted, battlefield, conveying a sense of sadness and despair., post-battle 

masterpiece 1902. the overall mood of the image is somber and melancholic,

This image is a captivating piece by the artist Osprey, known for their detailed and emotive style. It depicts a scene from the "Battle of Troy" series, specifically from "The Lords of Chaos" which involves two knights. This artwork beautifully captures the essence of medieval warfare and the somber atmosphere that is characteristic of Osprey's style.

best quality, newest 
\end{lstlisting} \\
\hline
\end{tabular}
\end{table}

\begin{table}[htbp]
\centering
\begin{tabular}{|>{\centering\arraybackslash}p{0.08\textwidth}|>{\centering\arraybackslash}p{0.86\textwidth}|}
\hline
\textbf{Image} & \includegraphics[width=0.85\textwidth,height=0.5\textheight,keepaspectratio]{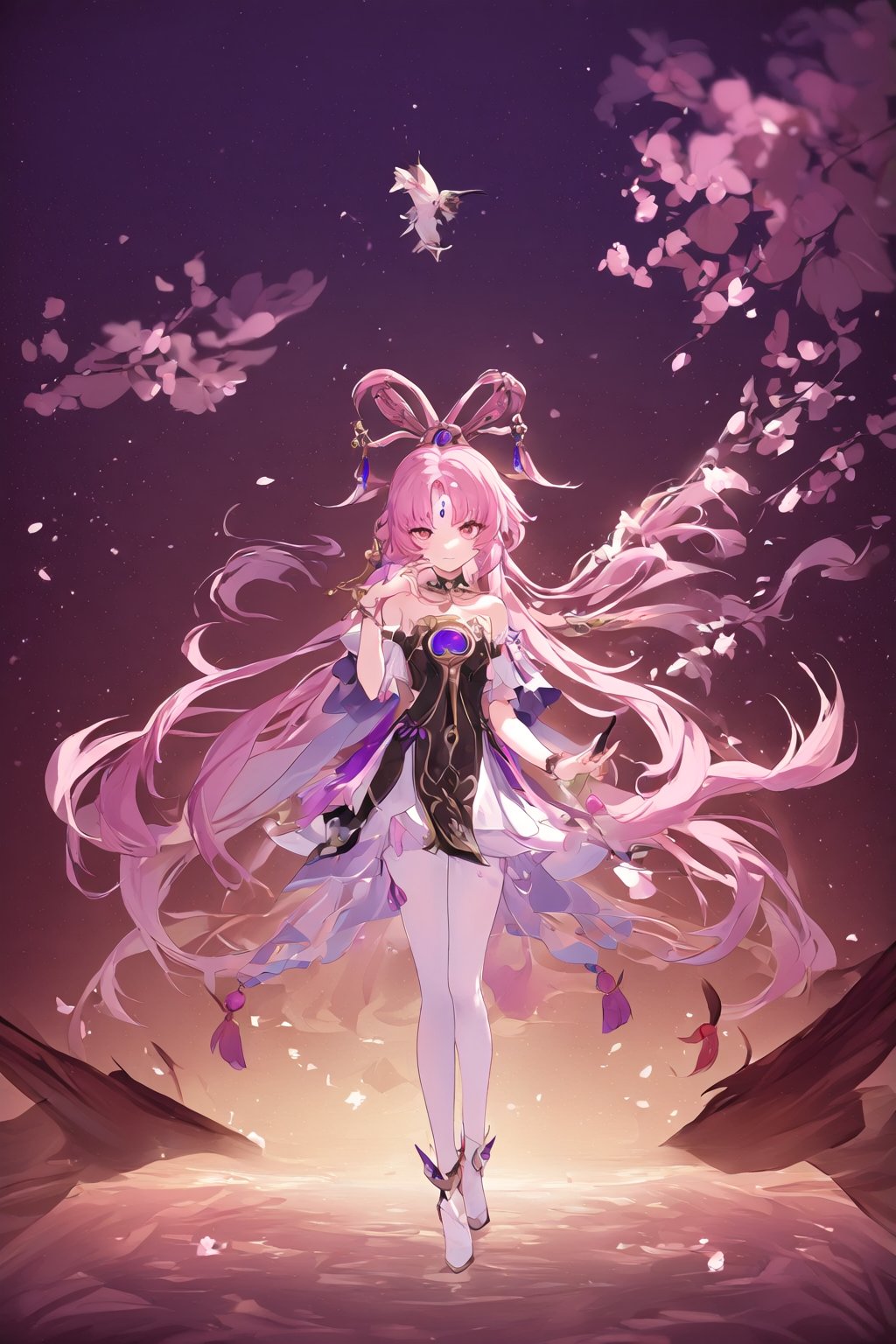} \\
\hline
\textbf{Prompt} & \begin{lstlisting}
1girl, fu xuan \(honkai: star rail\), honkai \(series\), honkai: star rail, yajuu, 

earrings, jewelry, arm up, tassel, solo, bird, pink hair, bare shoulders, parted bangs, detached sleeves, pink flower, hair ornament, cherry blossoms, holding, white pantyhose, looking at viewer, necklace, forehead jewel, long hair, very long hair, hair spread out, hair stick, short sleeves, pantyhose, floating hair, closed mouth, sidelocks, neck tassel, hair rings, flower, tassel hair ornament, dress, standing, white dre, pink eyes,

In this enchanting illustration by the artist yajuu, a character from the series "Honkai: Star Rail" stands amidst a vibrant display of cherry blossoms. The character, known as Fuyu Fu Xuan (Chinese), is depicted with long, flowing pink hair adorned with intricate hair ornaments and tassels. She wears traditional Chinese attire, including a dress with detached sleeves and white pantyhose, complemented by jewelry such as necklaces and earrings. Her eyes are closed, and she has a serene expression on her face, looking directly at the viewer. Fuyu Fu Xuan is surrounded by cherry blossom branches that burst with red flowers, creating a beautiful contrast against her pink attire. The background features a soft gradient of purple hues, further enhancing the ethereal atmosphere. A bird can be seen flying in the distance, adding to the sense of tranquility and harmony with nature. The overall composition is characterized by yajuu's signature style, which combines detailed line work with vivid colors and a dreamy ambiance.

masterpiece, newest, commentary request, absurdres 
\end{lstlisting} \\
\hline
\end{tabular}
\end{table}

\begin{table}[htbp]
\centering
\begin{tabular}{|>{\centering\arraybackslash}p{0.08\textwidth}|>{\centering\arraybackslash}p{0.86\textwidth}|}
\hline
\textbf{Image} & \includegraphics[width=0.85\textwidth,height=0.6\textheight,keepaspectratio]{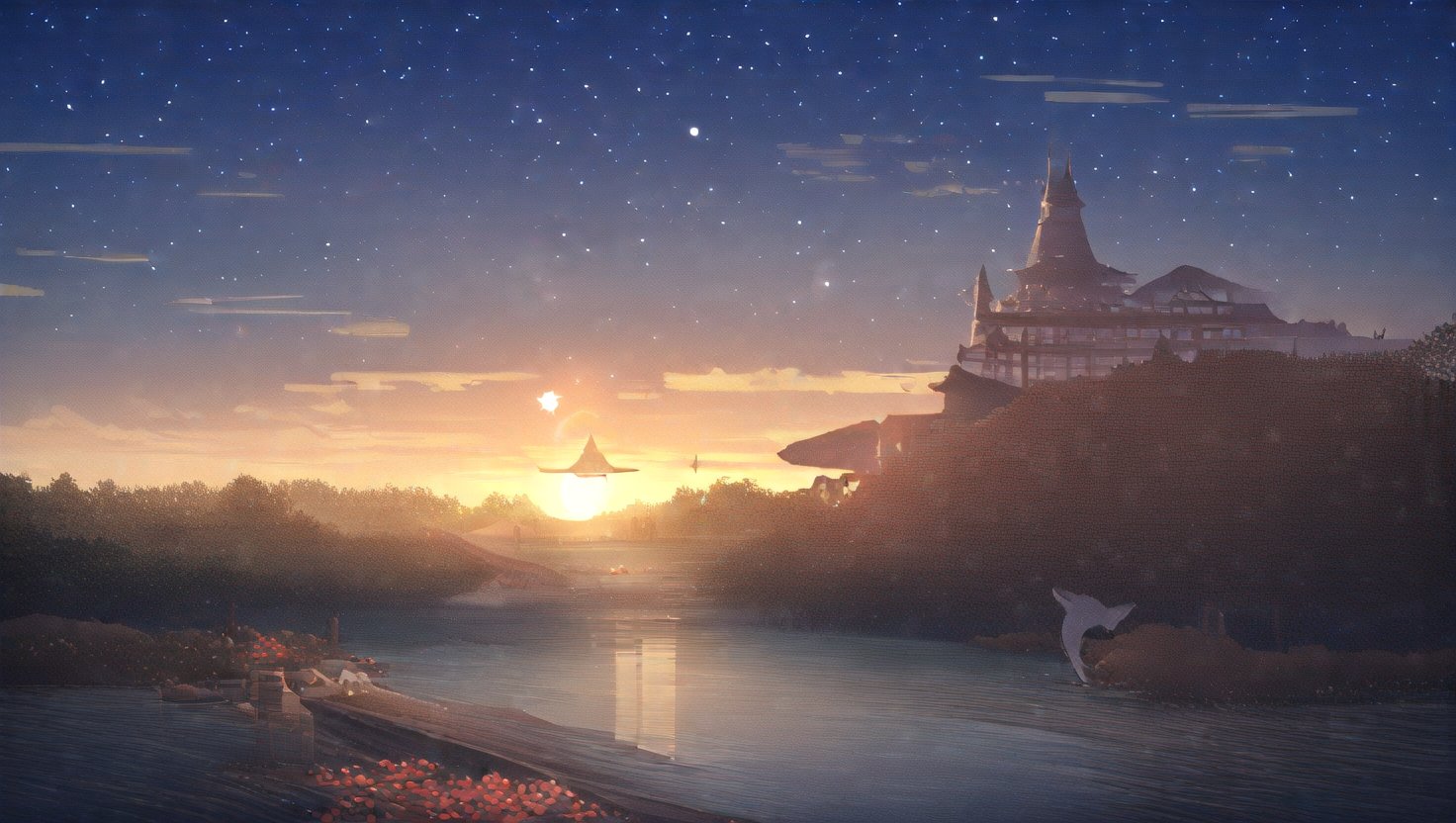} \\
\hline
\textbf{Prompt} & \begin{lstlisting}
maoqiu gungun lai berriiiq, scenery, best qualiyt, arai beta, arai beta maoqiu gungun lai berriiiq, sunset, ship, pond, pagoda, waterfall, landscape, watercraft, reflective water, moon, no humans, night sky, egasumi, tower, flower, lens flare, cloud, animal print, ryoubu torii, riverbank, forest, red scarf, night, blue theme, japanese flag, outdoors, boat, tree, bird, star \(symbol\), scarf, sunburst background, starry sky, oekaki, new year, crane \(machine\), kikumon, building, star \(sky\), nature, tiger lily, reflection, lake, white flower, blue scarf, full moon, a serene and vibrant digital illustration by Maoqiu GungunLai, characterized by its intricate detailing and rich color palette. The scene depicts a traditional Japanese bridge set against a tranquil landscape of mountains, trees with red flowers, and water. On the other side, a lone tiger, standing alert beside one of the riverbank where the blue sky above meets the calm water below, adds an element of adventure and intrigue to the composition. The artist's style is evident in the use of detailed patterns on various parts of the landscape, enhancing its overall beauty and complexity. masterpiece, great quality, sfw 
\end{lstlisting} \\
\hline
\end{tabular}
\end{table}

\begin{table}[htbp]
\centering
\begin{tabular}{|>{\centering\arraybackslash}p{0.08\textwidth}|>{\centering\arraybackslash}p{0.86\textwidth}|}
\hline
\textbf{Image} & \includegraphics[width=0.85\textwidth,height=0.6\textheight,keepaspectratio]{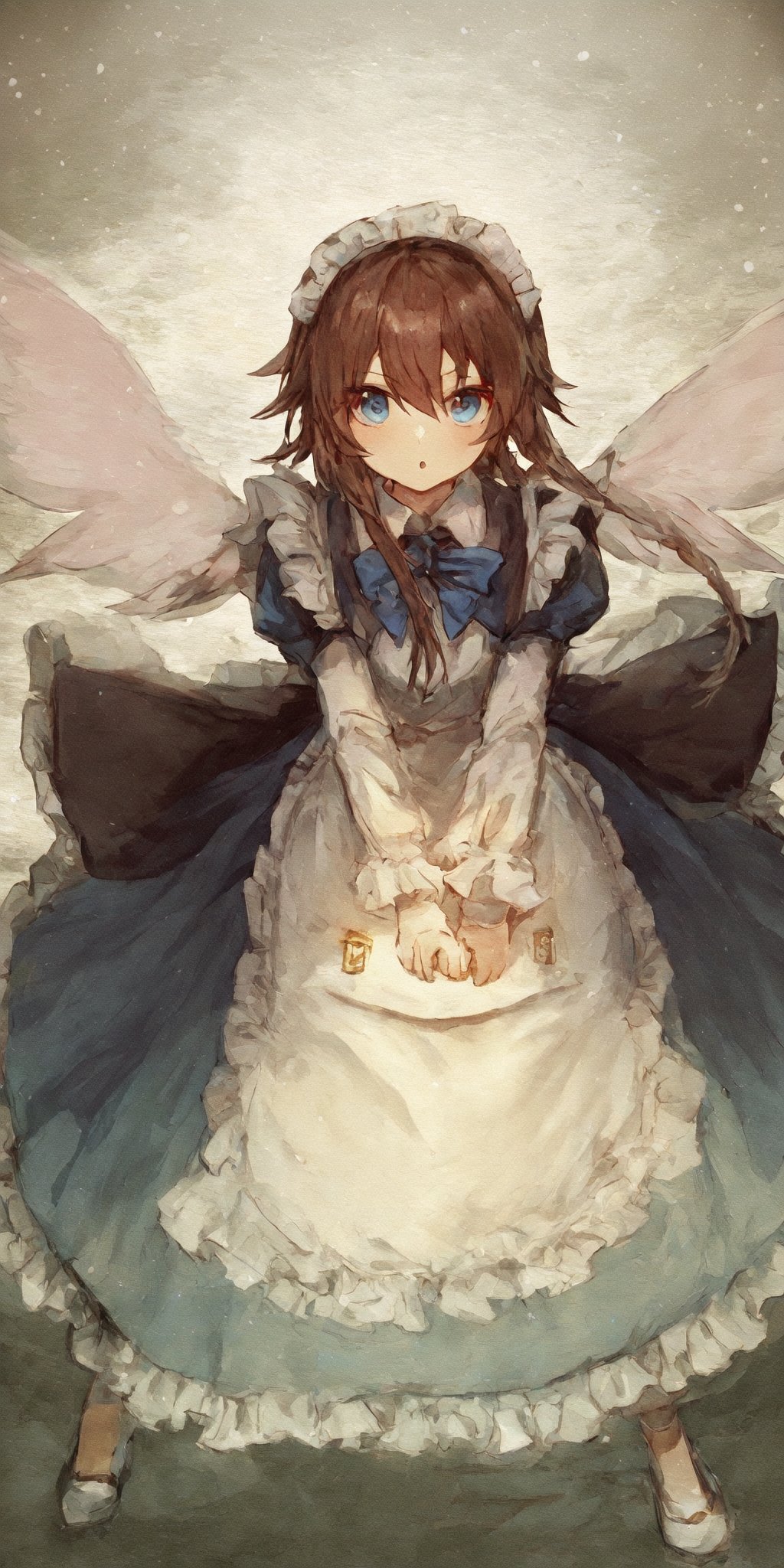} \\
\hline
\textbf{Prompt} & \begin{lstlisting}
1girl, 1girl, fairy maid \(touhou\), eguchi sera, touhou, saki, hi ros3, hi ros3,  scenery, best qualiyt, red eyes, blue eyes, short hair, brown hair, pants, frills, puffy sleeves, panties, blue dress, blue bowtie, open mouth, fairy wings, long sleeves, puffy long sleeves, german flag, frilled dress, sidelocks, bow, white footwear, fairy, full body, dress, apron, shirt, shoes, wings, collared shirt, juliet sleeves, short hair with long locks, bowtie, solo, hair between eyes, pink shirt, white collar, sign, holding, looking at viewer, maid apron, holding sign, pants under skirt, collar, maid headdress, fang, a captivating piece of digital art by hi ros3, characterized by its detailed and vibrant style. The central figure is an archer from the Touhou series, distinguished by her long brown hair with sidelocks and striking blue eyes. Her pose and attire suggest a sense of grace and readiness, capturing a moment of poise within this fantastical world. masterpiece, great quality, sfw Negative prompt: unknownlow quality, worst quality, text, signature, jpeg artifacts, bad anatomy, old, early, copyright name, watermark, artist name, signature, weibo username, mosaic censoring, bar censor, censored, text, speech bubbles, doll, character doll, hair intake, realistic, 2girls, 3girls, multiple girls, crop top, cropped head, cropped, large breasts, nsfw Steps: 35, Sampler: euler_simple, CFG Scale: 4.0, Seed: 291795535432825, Size: 1024x2048, Model hash: f7e7841239, Model: hdm-xut-340M-1024px-9kstep-note, Version: ComfyU 
\end{lstlisting} \\
\hline
\end{tabular}
\end{table}